\documentclass[sigconf]{acmart}
\renewcommand\footnotetextcopyrightpermission[1]{}

\usepackage{colortbl}
\usepackage{booktabs} 
\usepackage{algorithm}
\usepackage{amsmath}
\usepackage{graphicx}
\usepackage{amsfonts,bm}
\usepackage{dcolumn}
\usepackage{multirow}
\usepackage{color}
\usepackage{braket}
\usepackage{caption}
\usepackage{subcaption}
\usepackage{mathtools}
\usepackage{enumerate}
\usepackage{enumitem}
\usepackage{empheq}
\usepackage{calc}
\usepackage{dsfont}
\usepackage{algpseudocode}
\usepackage{enumitem}
\usepackage{makecell}
\usepackage{geometry} 
\usepackage{xspace}
\AtBeginDocument{%
  }

\newcommand{\std}[1]{{\scriptstyle \pm #1}}
\newcommand{\mthd}{RACI\xspace}

\acmConference[Conference acronym 'XX]{Make sure to enter the correct
  conference title from your rights confirmation email}{June 03--05,
  2018}{Woodstock, NY}

\begin{document}

\title{Role-Aware Conditional Inference for Spatiotemporal Ecosystem Carbon Flux Prediction}

\author{Yiming Sun}
\affiliation{%
  \institution{University of Pittsburgh}
  \city{Pittsburgh}
  \state{PA}
  \country{USA}
}
\email{yimingsun@pitt.edu}

\author{Runlong Yu}
\affiliation{%
  \institution{University of Alabama}
  \city{Tuscaloosa}
  \state{AL}
  \country{USA}
}
\email{ryu5@ua.edu}

\author{Rongchao Dong}
\affiliation{%
  \institution{University of Pittsburgh}
  \city{Pittsburgh}
  \state{PA}
  \country{USA}
}
\email{rongchaodong@pitt.edu}

\author{Shuo Chen}
\affiliation{%
  \institution{Purdue University}
  \city{West Lafayette}
  \state{IN}
  \country{USA}
}
\email{chen4371@purdue.edu}

\author{Licheng Liu}
\affiliation{%
  \institution{University of Wisconsin-Madison}
  \city{Madison}
  \state{WI}
  \country{USA}
}
\email{licheng.liu@wisc.edu}

\author{Youmi Oh}
\affiliation{%
  \institution{NOAA Global Monitoring Laboratory}
  \city{Boulder}
  \state{CO}
  \country{USA}
}
\email{youmi.oh@noaa.gov}

\author{Qianlai Zhuang}
\affiliation{%
  \institution{Purdue University}
  \city{West Lafayette}
  \state{IN}
  \country{USA}
}
\email{qzhuang@purdue.edu}

\author{Yiqun Xie}
\affiliation{%
  \institution{University of Maryland}
  \city{College Park}
  \state{MD}
  \country{USA}
}
\email{xie@umd.edu}

\author{Xiaowei Jia}
\affiliation{%
  \institution{University of Pittsburgh}
  \city{Pittsburgh}
  \state{PA}
  \country{USA}
}
\email{xiaowei@pitt.edu}

\renewcommand{\shortauthors}{Sun et al.}
\setlength{\abovedisplayskip}{3pt} 
\setlength{\belowdisplayskip}{3pt} 
\setlength{\abovedisplayshortskip}{0pt} 
\setlength{\belowdisplayshortskip}{3pt}
\setlist[itemize]{topsep=2pt, itemsep=1.5pt, parsep=0pt}

\begin{abstract}
Accurate prediction of terrestrial ecosystem carbon fluxes (e.g., CO$_2$, GPP, and CH$_4$) is essential for understanding the global carbon cycle and managing its impacts. However, prediction remains challenging due to strong spatiotemporal heterogeneity: ecosystem flux responses are constrained by slowly varying regime conditions, while short-term fluctuations are driven by high-frequency dynamic forcings. Most existing learning-based approaches treat environmental covariates as a homogeneous input space, implicitly assuming a global response function, which leads to brittle generalization across heterogeneous ecosystems. In this work, we propose \textit{Role-Aware Conditional Inference} (RACI), a process-informed learning framework that formulates ecosystem flux prediction as a conditional inference problem. RACI employs hierarchical temporal encoding to disentangle slow regime conditioners from fast dynamic drivers, and incorporates role-aware spatial retrieval that supplies functionally similar and geographically local context for each role. By explicitly modeling these distinct functional roles, RACI enables a model to adapt its predictions across diverse environmental regimes without training separate local models or relying on fixed spatial structures. We evaluate RACI across multiple ecosystem types (wetlands and agricultural systems), carbon fluxes (CO$_2$, GPP, CH$_4$), and data sources, including both process-based simulations and observational measurements. Across all settings, RACI consistently outperforms competitive spatiotemporal baselines, demonstrating improved accuracy and spatial generalization under pronounced environmental heterogeneity.
\end{abstract}

\maketitle

\section{Introduction}
Terrestrial ecosystems play an indispensable role in the global carbon budget by regulating the exchange of greenhouse gases between the land surface and the atmosphere through the cycling of carbon dioxide and methane~\cite{reichstein2013climate,saunois2019global}. Accurately monitoring and modeling these fluxes is critical for understanding carbon-climate feedbacks and informing effective mitigation strategies~\cite{jung2020scaling,tramontana2016predicting}. 
However, learning globally consistent flux–driver relationships remains challenging because the underlying processes are highly heterogeneous. In addition, sparse observations force models to extrapolate from a few sites to vast, biophysically distinct regions, degrading location-specific performance~\cite{jung2020scaling,tian2021impacts, zhang2025ensemble}.

Process-based models (PBMs) have long been a dominant paradigm for terrestrial carbon flux modeling, using process-level equations grounded in physical and biological principles to simulate biogeochemical dynamics~\cite{CLM,ORCHIDEE,del2005daycent}. While offering interpretability and physical consistency, PBMs incur high computational costs and rely on fixed model structures and semi-empirical parameterizations that are often calibrated using limited site- or vegetation-specific observations, which substantially limits their transferability across diverse global ecosystems~\cite{reichstein2013climate,reichstein2019deep}.
In parallel, machine learning (ML)-based approaches have gained increasing attention for their ability to capture complex nonlinear relationships from data, and can match or exceed PBMs when sufficient observations are available~\cite{dou2018comprehensive,guevara2021machine,tramontana2020partitioning,chen2024quantifying}. To mitigate the impact of sparse observations, recent hybrid approaches integrate domain knowledge from PBMs into learning-based frameworks, aiming to improve robustness and extrapolation under data-scarce conditions~\cite{liu2024knowledge,cheng2025knowledge,rong2025hybrid}.

Despite these advances, most existing ML methods still rely on a homogeneous modeling assumption, learning a single global mapping from environmental variables to carbon fluxes~\cite{jung2020scaling,reitz2021upscaling,chen2024quantifying}. Under strong ecosystem heterogeneity and distribution shifts, such global models tend to converge toward 
global averages~\cite{ploton2020spatial,meyer2022machine,ludwig2024resolving}, leading to degraded location-specific  predictions and poor generalization to unmonitored regions or unseen weather conditions~\cite{meyer2022machine,beucler2024climate}. 
In particular, we identify two key challenges in explicitly modeling ecosystem heterogeneity for carbon prediction.



\begin{figure}[h]
\centering
\includegraphics[width=\linewidth]{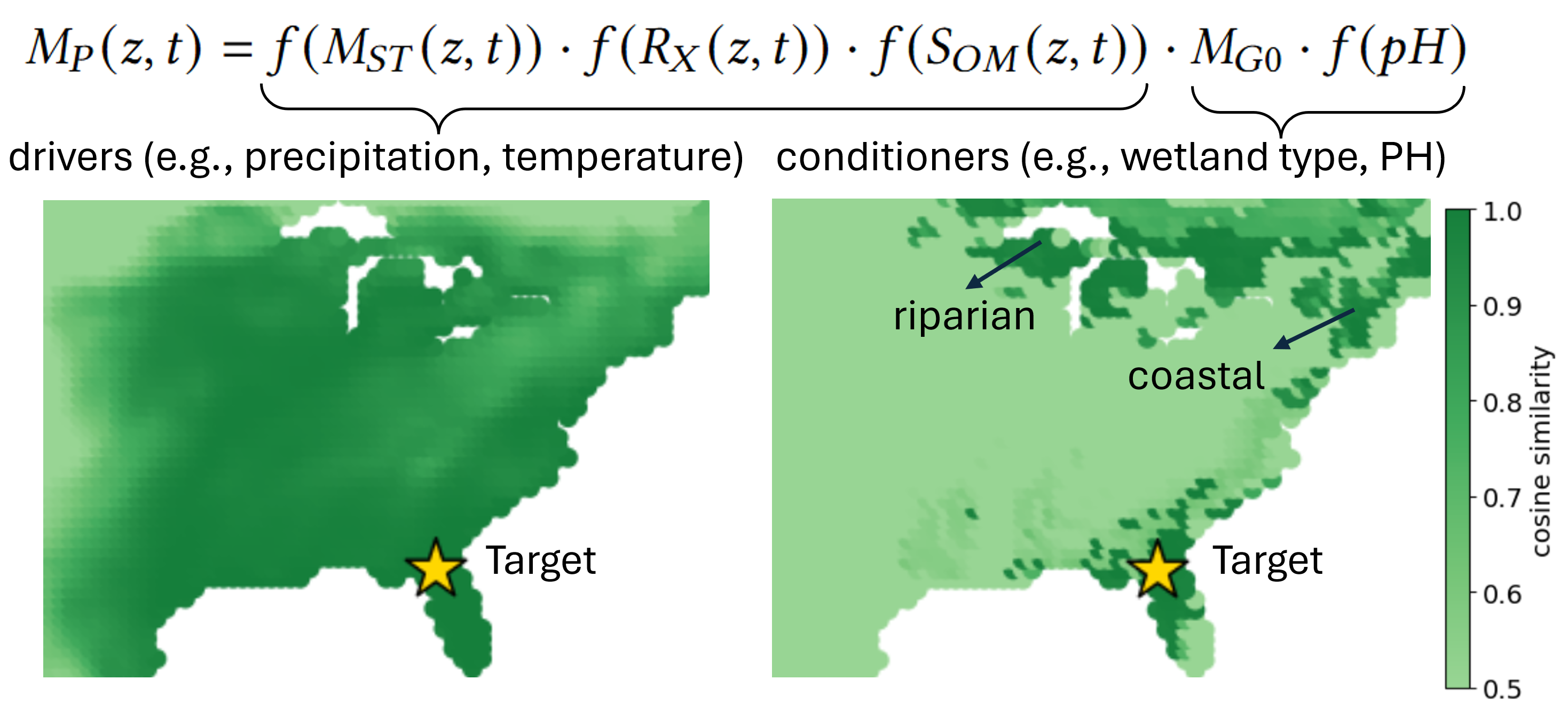}
\vspace{-0.15in}
\caption{Illustration of (i) functional heterogeneity between fast drivers and slow conditioners, and (ii) their contrasting spatial similarity patterns.}
\vspace{-0.15in}
\label{fig:heterogeneity_illustration}
\end{figure}

The first challenge arises from the functional heterogeneity of environmental variables involved in ecosystem processes. 
As illustrated by the TEM-MDM~\cite{TEM} formulation in Figure~\ref{fig:heterogeneity_illustration}, different variables influence carbon fluxes through distinct mechanisms and characteristic time scales~\cite{carvalhais2014global} (detailed in Appendix~\ref{sec:process_appendix}). 
Specifically, empirical evidence and PBMs indicate that some variables describe slowly evolving background conditions, such as soil properties or long-term climate states, that condition the form and magnitude of ecosystem responses over extended periods~\cite{besnard2019memory}. In contrast, fast-varying drivers such as temperature and precipitation induce short-term fluctuations in carbon fluxes~\cite{reichstein2013climate,besnard2019memory}. 
However, standard ML models often treat all environmental variables as homogeneous covariates in a single response function
and do not explicitly separate slowly varying background conditioners from fast-varying dynamic drivers, or represent temporal patterns at multiple scales~\cite{reichstein2019deep,xiao2012advances,besnard2019memory}. 


The second challenge lies in modeling spatial heterogeneity under limited observations.
Ecosystem data vary substantially across space, reflecting differences in both environmental variables and flux responses~\cite{li2022spatial}. This heterogeneity is further complicated by the distinct spatial distributions of fast drivers versus slow background conditioners. Fast meteorological drivers such as precipitation and temperature tend to vary smoothly across space and are physically coupled across neighboring sites through lateral thermal, hydrological and biogeochemical processes, so that local states such as water table depth and/or soil organic matter are strongly influenced by surrounding landscapes~\cite{sakaguchi2022determining,saunois2019global}. In contrast, slow background conditioners such as climate, land covers, and soil properties often form fragmented, mosaic-like patterns, so geographically distant sites can share similar biophysical regimes and response behaviors~\cite{negassa2022spatial,li2022spatial}. As illustrated in Figure~\ref{fig:heterogeneity_illustration}, we take a site in Florida and compute the cosine similarity between its feature representations and those of other grid cells across North America to visualize this pattern. This combination produces a mixture of gradual and abrupt shifts across regions, complicating the learning of spatially consistent flux–driver relationships.
Given the sparse and uneven distribution of flux observations, it is infeasible to train specialized models for each regime, making it essential to develop approaches that can adapt predictions to diverse spatial contexts without dense local supervision.

To tackle these challenges, we propose \textit{Role-Aware Conditional Inference} (\mthd), a process-informed learning framework that aligns model structure with role-dependent ecosystem processes. To address functional heterogeneity, \mthd employs a Role-Separating Temporal Modeling design, which explicitly decouples slowly evolving background conditioners from fast-varying dynamic drivers, so that short-term predictions are conditioned on regime context. To handle spatial heterogeneity, \mthd incorporates a Role-Aware Spatial Contextual Retrieval module tailored to the spatial structure of each role: it retrieves regime-similar context for background conditioners while aggregating localized forcing patterns for dynamic drivers.
Together, these components allow a single, unified predictor to adapt its response logic across diverse biophysical regimes, avoiding fragmented local models and enabling robust, scalable carbon flux prediction in data-sparse regions.

We evaluate \mthd across a diverse set of ecosystem modeling scenarios to assess its robustness and generalizability.
Our experiments span multiple land systems (e.g., agricultural regions and wetlands), carbon-related fluxes (CO$_2$, GPP, and CH$_4$), and multiple data sources, including both process-based simulations and sparse observational measurements.
Across these settings, \mthd consistently outperforms a broad set of competitive baselines, demonstrating that its performance gains are not confined to specific ecosystem types or data modalities, but rather stem from its fundamental ability to reconcile role-dependent process structures.

In summary, our main contributions are:
\begin{itemize}[leftmargin=1.5em]
    \item We formalize functional heterogeneity in terrestrial ecosystem modeling by distinguishing between background conditioners and dynamic drivers, highlighting the limitations of standard homogeneous global models.
    \item We propose a Role-Aware Conditional Inference (\mthd) framework that aligns model architecture with these roles through role-separating temporal modeling and role-aware spatial retrieval, enabling adaptive, context-sensitive predictions.
    \item We evaluate \mthd across diverse ecosystem types, carbon fluxes, and data sources, where it consistently outperforms competitive baselines and improves global-scale carbon cycle modeling under sparse observations.
\end{itemize}



\section{Preliminaries}
\label{sec:problem_definition}

\begin{figure*}[h]
\centering
\vspace{-0.1in}
\includegraphics[width=\linewidth]{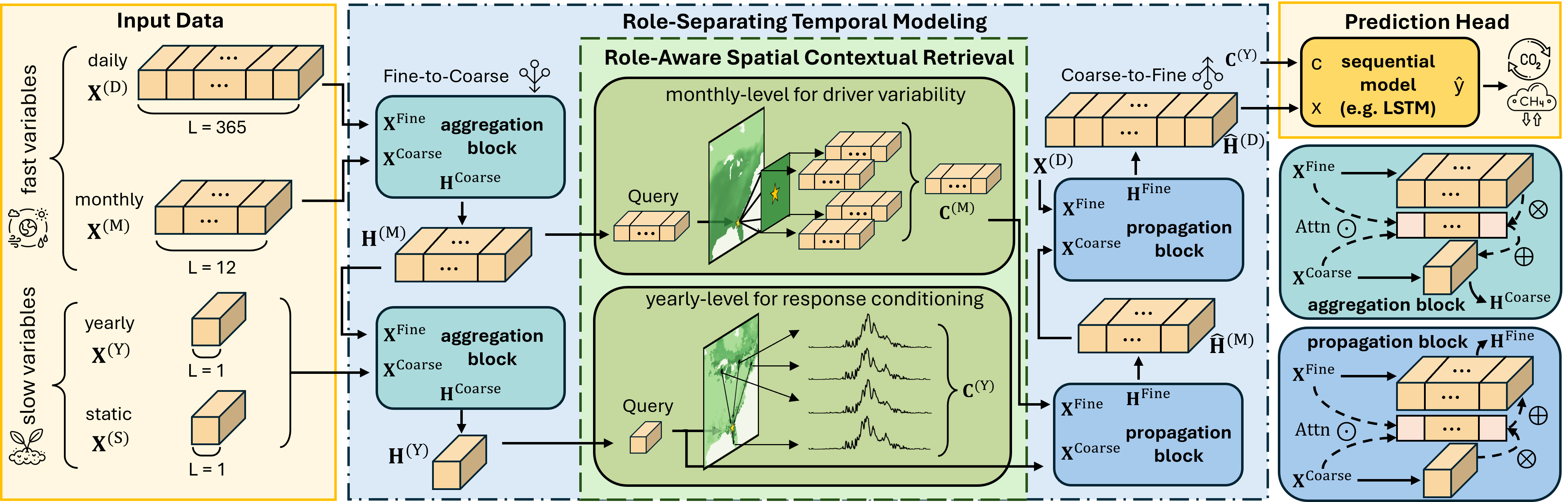}
\vspace{-0.1in}
\caption{Overview of the \mthd framework, with two coupled components: Role-Separating Temporal Modeling (blue) and Role-Aware Spatial Contextual Retrieval (green). The right panel details the cross-scale attention blocks.}
\vspace{-0.05in}
\label{fig:framework}
\end{figure*}

We study a spatiotemporal prediction problem over a set of monitoring sites $\mathcal{S}$, where each sample represents a \emph{site-year} pair consisting of hierarchical observations and daily ecosystem flux targets.

\paragraph{Hierarchical Input Structure.}
The input features arise from distinct generative processes with mismatched timescales. For a site $s$ in year $y$, the input $\mathbf{X}_{s,y} = \{\mathbf{X}_{s,y}^{(D)}, \mathbf{X}_{s,y}^{(M)}, \mathbf{X}_{s,y}^{(Y)}, \mathbf{X}^{(S)}_s\}$ comprises:

\begin{itemize}[leftmargin=1em]
    \item \textbf{Daily drivers} $\mathbf{X}_{s,y}^{(D)} \in \mathbb{R}^{365 \times d_D}$: Capturing high-frequency meteorological forcing (e.g., precipitation, temperature) that directly induces short-term variability in ecosystem fluxes.
    
    \item \textbf{Monthly drivers} $\mathbf{X}_{s,y}^{(M)} \in \mathbb{R}^{12 \times d_M}$: Representing seasonal ecosystem information  (e.g., net primary productivity).
    
    \item \textbf{Yearly trends} $\mathbf{X}_{s,y}^{(Y)} \in \mathbb{R}^{1 \times d_Y}$: Characterizing slowly varying background trends (e.g., global CO$_2$ concentration).
    
    \item \textbf{Static attributes} $\mathbf{X}^{(S)}_s \in \mathbb{R}^{d_S}$: Describing time-invariant structural properties of the site (e.g., soil texture, climate zone).
\end{itemize}


\paragraph{Learning Objective.}
We aim to learn a mapping $\mathcal{F}$ such that $\hat{\mathbf{Y}}_{s,y} = \mathcal{F}(\mathbf{X}_{s,y})$ predicts the daily flux sequence $\mathbf{Y}_{s,y} \in \mathbb{R}^{365}$. The optimal parameters are learned by minimizing
$\sum_{(s,y) \in \mathcal{D}_\text{train}} \mathcal{L}(\mathbf{Y}_{s,y}, \hat{\mathbf{Y}}_{s,y}),$
where $\mathcal{L}$ is a loss function (e.g., Mean Squared Error).

\paragraph{Spatiotemporal Distribution Shift.}
To formalize distributional shift, we decompose the joint distribution of inputs $\mathbf{X}$ and targets $\mathbf{Y}$ as $p(\mathbf{X}, \mathbf{Y}) = p(\mathbf{Y}|\mathbf{X})p(\mathbf{X})$ and spatiotemporal heterogeneity manifests as two distinct forms of distribution shift:

\begin{itemize}[leftmargin=15pt]
    \item \textbf{Forcing shift $p(\mathbf{X})$:} Variations in the input distribution of dynamic drivers due to regional differences in weather conditions. 
    
    \item \textbf{Response shift $p(\mathbf{Y}|\mathbf{X})$:} Non-stationarity in the mapping from drivers to fluxes, governed by biophysical regimes.
\end{itemize}

The coexistence of these shifts implies that a universal predictor trained on sparse observations is prone to bias, as it tends to learn a global average that fails to capture heterogeneous ecosystems.

\section{Methodology}

\subsection{The \mthd Framework}
\label{sec:problem_formulation}

To address the spatiotemporal distribution shift discussed in Section~\ref{sec:problem_definition}, we reformulate the task as a conditional inference problem: $f\colon(\textbf{X}, \textbf{C}) \rightarrow \textbf{Y}$, where $\textbf{X}$ denotes local site-year inputs and $\textbf{C}$ provides auxiliary contextual information that adapts the model to the underlying environmental regime. Inspired by in-context learning, this context $\textbf{C}$ serves as a data-driven prior that informs the model about which regime it is in and allows a single predictor to adjust its behavior across heterogeneous ecosystems without site-specific retraining.

We introduce \textbf{\mthd}, a process-informed framework designed to instantiate this conditional inference through two coupled components, \textit{Role-Separating Temporal Modeling} and \textit{Role-Aware Spatial Contextual Retrieval} (Figure~\ref{fig:framework}). Together, these modules align model behavior with role- and scale-dependent ecosystem processes.

\textbf{Role-Separating Temporal Modeling.}
To structure hierarchical inputs by functional role and characteristic timescale, \mthd decomposes input $\mathbf{X}$ into conditioners and drivers. Conditioners, such as static soil properties and annual trends, shape the response mechanism $p(\mathbf{Y}|\mathbf{X})$ by defining the long-term regime in which the ecosystem operates. High-frequency drivers govern short-term variability and the input distribution $p(\mathbf{X})$. By separating these roles, the model interprets dynamic drivers conditioned on explicit representations of background conditioners, rather than implicitly averaging over incompatible biophysical regimes. 

\textbf{Role-Aware Spatial Contextual Retrieval.}
To address distribution shifts explicitly, we construct the context $\mathbf{C} = \{\mathbf{C}^{(M)}, \mathbf{C}^{(Y)}\}$ via scale-dependent retrieval. To capture regional seasonal weather information $\mathbf{C}^{(M)}$, we perform retrieval over spatially adjacent sites and aggregate their signals at the monthly scale. 
Here we avoid daily-level retrieval because daily drivers contain frequent fluctuations that can obscure the underlying seasonal signal. 
The obtained monthly context is then used to contextualize local drivers and mitigate the forcing shift in $p(\mathbf{X})$. In parallel, we address response shift in $p(\mathbf{Y}\mid\mathbf{X})$ via yearly-level global retrieval: we find site–years with similar biophysical conditioners and aggregate their flux trajectories into a functional prior $\mathbf{C}^{(Y)}$ that guides the response for the target site-year. The resulting learning objective minimizes $\mathcal{L}\big(\mathbf{Y}, f(\mathbf{X}, \mathbf{C}^{(Y)}, \mathbf{C}^{(M)})\big)$, making predictions explicitly conditioned on both local inputs and retrieved spatial context, and supporting robust generalization under environmental heterogeneity.

\subsection{Role-Separating Temporal Modeling}

The temporal component of \mthd organizes the hierarchical inputs into three aligned pathways: high-frequency drivers at daily scale, seasonal drivers at monthly scale, and slow-varying conditioners at yearly scale. The goal is to obtain consistent representations where fine-scale embeddings primarily reflect the input distribution $p(\mathbf{X})$, while coarse-scale embeddings summarize regime information that shapes the conditional response $p(\mathbf{Y}|\mathbf{X})$. These role-consistent representations form the basis for subsequent spatial retrieval and conditional inference.
To achieve this, \mthd adopts a hierarchical encoder that couples daily, monthly, and yearly levels through selective cross-scale interactions. Information is first aggregated from fine to coarse scales to form stable regime summaries, and then propagated from coarse to fine scales so that high-frequency drivers are 
modulated by the appropriate regime context. This targeted, asymmetric information flow yields role-consistent temporal representations and naturally separates slow-varying conditioners from dynamic drivers in a physically consistent manner. 

\subsubsection{Fine-to-coarse aggregation.}

In the fine-to-coarse direction (i.e., daily $\rightarrow$ monthly $\rightarrow$ yearly), temporal aggregation aims to build coarse-scale representations that remain informative for the subsequent retrieval process. 
Instead of using fixed averaging, we employ a data-driven aggregation mechanism that learns which fine-scale patterns should be retained in the coarser-scale representations. 
For example, we expect such aggregation can better represent extreme events that may occur only a few times in a year, so the resulting representation distinguishes a year with rare extremes from a normal year with sustained moderate conditions, even when their monthly or yearly summary statistics appear similar. 
By assigning higher weights to informative periods and down-weighting less relevant patterns, the aggregated representations effectively summarize the forcing trajectory under which the ecosystem operated. 

We apply this aggregation mechanism hierarchically, first from the daily to the monthly scale and then to the yearly scale. Here we illustrate how the yearly embedding $\mathbf{H}^{(Y)}_{s,y}$ is constructed by aggregating monthly embeddings $\{\mathbf{H}_{s,y,m}^{(M)}\}_{m=1}^{12}$, which themselves are obtained by aggregating daily representations. At the yearly level, we first concatenate yearly trends $\mathbf{X}_{s,y}^{(Y)}$ and static attributes $\mathbf{X}_{s,y}^{(S)}$ into a joint regime vector $\mathbf{X}_{s,y}^{(R)} = [\mathbf{X}_{s,y}^{(Y)}; \mathbf{X}_{s,y}^{(S)}]$. We then use this  vector  as the attention query over the monthly embeddings and as a residual component to form the final yearly embedding:
\begin{equation}
\begin{aligned}
\mathbf{H}^{(Y)}_{s,y} &= \sum_{m=1}^{12} \alpha_{s,y,m} \mathbf{H}^{(M)}_{s,y,m} + \phi_R(\mathbf{X}^{(R)}_{s,y}), \\
\alpha_{s,y,m} &= \operatorname{Attn}(\phi_R(\mathbf{X}^{(R)}_{s,y}), \mathbf{H}^{(M)}_{s,y,m}),
\end{aligned}
\end{equation}
where $\phi_R$ is a yearly encoder applied to $\mathbf{X}^{(R)}_{s,y}$, and $\operatorname{Attn}$ denotes dot-product attention that produces normalized weights $\alpha_{s,y,m}$ over months $m$ for site $s$ in year $y$.
In this way, $\mathbf{H}^{(Y)}_{s,y}$ integrates an attention-weighted summary of intra-annual dynamics with the yearly background and static site properties to form a compact regime descriptor. The daily-to-monthly aggregation follows an analogous query-key mechanism.

\subsubsection{Coarse-to-fine propagation.}
In the coarse-to-fine direction (i.e., yearly $\rightarrow$ monthly $\rightarrow$ daily), temporal propagation specifies how coarse-scale information modulates fine-scale dynamics. A naive choice is to simply replicate coarse embeddings across all finer time steps, implicitly assuming a piecewise-constant influence over time. In practice, however, regime-level conditions may have varying influence on the flux dynamics across months or days. 

To capture this variability, \mthd propagates coarse information to finer scales through a learned gating mechanism. Specifically, for each month $m$, we compute a scalar gate $\beta_{s,y,m}$ that measures the compatibility between the yearly regime and the seasonal embedding, and use it to inject regime information into the monthly representation, which generates a refined version of the original monthly embedding:
\begin{equation}
\begin{aligned}
\tilde{\mathbf{H}}^{(M)}_{s,y,m} &= \mathbf{H}^{(M)}_{s,y,m} + \beta_{s,y,m} \cdot \mathbf{H}^{(Y)}_{s,y}, \\
\quad \beta_{s,y,m} &=  \operatorname{GateAttn}(\mathbf{H}^{(Y)}_{s,y}, \mathbf{H}^{(M)}_{s,y,m}).
\end{aligned}
\end{equation}
where $\operatorname{GateAttn}$ is a cross-attention network that outputs an unnormalized relevance score $\beta_{s,y,m}$ for each month. Unlike standard attention, these gates are not constrained to sum to one across $m$, allowing the model to amplify or suppress the influence of the yearly regime at different times. We follow a similar process to further propagate monthly information from $\tilde{\mathbf{H}}^{(M)}$ into daily embeddings. This propagation ensures that the resulting daily embeddings  $\tilde{\mathbf{H}}^{(D)}$ are informed by the annual regime while preserving the heterogeneous variance of the input distribution $p(\mathbf{X})$.

\subsection{Role-Aware Spatial Contextual Retrieval}

To connect the distribution-shift perspective with downstream prediction, \mthd injects role-specific spatial context derived from both the input distribution $p(\mathbf{X})$ and the conditional response $p(\mathbf{Y}|\mathbf{X})$ into the flux prediction task. Concretely, it constructs two contextual signals: a driver-oriented context that summarizes regional forcing patterns and a conditioner-oriented context that captures regime-similar flux behavior, which together form the functional prior used in our conditional inference framework.

\subsubsection{Monthly-level retrieval for driver variability.}

At finer temporal scales, the dominant challenge is the spatial heterogeneity in environmental drivers $p(\mathbf{X})$. Although one could retrieve neighbors at the daily level, daily meteorological inputs (e.g., transient precipitation or localized cloud cover) are often dominated by high-frequency variability. 
Such "weather-level fluctuations" capture variability beyond the site characteristics that drive spatial shifts in $p(\mathbf{X})$ and thus are not suitable for robust spatial conditioning.
We therefore perform retrieval at the monthly scale, where aggregation suppresses day-to-day noise and yields a more stable regional weather signature. 
For a target site $s$ in year $y$ and month $m$, we retrieve a neighborhood of geographically adjacent sites $\mathcal{N}(s)$ and aggregate their monthly embeddings into a driver-oriented spatial context:
$\mathbf{C}^{(M)}_{s,y,m} = \sum_{s' \in \mathcal{N}(s)} \operatorname{Attn}(\mathbf{H}^{(M)}_{s,y,m}, \mathbf{H}^{(M)}_{s',y,m}) \cdot \mathbf{H}^{(M)}_{s',y,m}.$
This 
context information is then propagated coarse-to-fine to the daily level, ensuring that fine-grained predictions are grounded in a stable regional forcing context.

\subsubsection{Yearly-level retrieval for response conditioning.}

While monthly drivers are often spatially smooth, the underlying biophysical mechanisms  $p(\mathbf{Y}|\mathbf{X})$ can be highly heterogeneous and typically follow a mosaic-like spatial pattern. Sites with similar weather drivers may exhibit very different flux responses due to differences in soil organic matter, vegetation types, or microbial communities. The response shift means that geographic proximity is often a poor proxy for functional similarity.
To address this, we perform global retrieval at the yearly scale to find functional analogues. To avoid temporal leakage, retrieval is restricted to a held-out auxiliary pool $\mathcal{A}$ of site-years that is disjoint from both the training and test splits and excludes the target prediction year.

Retrieval is formulated as a cross-attention operation where the target yearly embedding acts as the query and auxiliary site-years provide keys and values. For a target site-year $(s,y)$ with embedding $\mathbf{H}^{(Y)}_{s,y}$, we first identify a candidate set $\mathcal{R}(s,y) \subset \mathcal{A}$ by retaining site-years whose similarity to $\mathbf{H}^{(Y)}_{s,y}$ exceeds a threshold $\tau$. This step removes clearly mismatched regimes and focuses on functionally plausible analogues. The yearly context is then constructed by attention-based aggregation of normalized flux trajectories: 
$$
\mathbf{C}^{(Y)}_{s,y} = \sum_{(s',y') \in \mathcal{R}(s,y)} \text{Attn}(\mathbf{H}^{(Y)}_{s,y}, \mathbf{H}^{(Y)}_{s',y'}) \cdot \tilde{\mathbf{Y}}_{s',y'},
$$
where $\tilde{\mathbf{Y}}_{s',y'}$ denotes the magnitude-normalized flux trajectory for site-year $(s',y')$. 
By focusing on normalized trajectories, the retrieval emphasizes the temporal response pattern—the characteristic "shape" of how a specific ecosystem reacts to forcing—rather than absolute flux magnitudes that may be affected by local scaling factor.
If no auxiliary site-year satisfies the similarity threshold (i.e., $\mathcal{R}(s,y) = \emptyset$), we set $\mathbf{C}^{(Y)}_{s,y} = \mathbf{0}$ and skip response conditioning for that site-year, so that the model falls back to using only local information rather than forcing mismatched contextual trajectories.

Because $\mathbf{H}^{(Y)}_{s,y}$ encodes  
both yearly background conditions and aggregated forcing for the target site-year, retrieval is effectively performed in a regime space. The selected trajectories thus represent response patterns that are functionally consistent with the target ecosystem, allowing the model to “borrow’’ response behaviors from globally similar site-years. The daily decoder then combines this retrieved context with local inputs to instantiate an appropriate response for the target, resolving functional heterogeneity in a regime-aware manner without introducing site-specific parameters or training separate local models.

\subsection{Prediction and Optimization}

Given the role-separated temporally encoded representations and role-aware retrieved spatial contexts, \mthd generates daily flux predictions through a response-conditioning prediction head. This stage combines retrieved response priors within the specific local environmental context, effectively transforming the model from a static global predictor into a regime-adaptive predictor.
Specifically, the LSTM-based prediction head treats the retrieved yearly context $\mathbf{C}^{(Y)}_{s,y}$ as a mechanistic prior that guides how daily representations $\tilde{\mathbf{H}}^{(D)}_{s,y}$ are interpreted:
$$
\hat{\mathbf{Y}}_{s,y}
=
\mathrm{LSTM}
\left(
\tilde{\mathbf{H}}^{(D)}_{s,y}
\,\|\, 
\mathbf{C}^{(Y)}_{s,y}
\right).
$$

The entire framework is trained end-to-end by minimizing error between predicted and observed daily flux sequences.

\section{Experiments}

We conduct experiments to evaluate the effectiveness of \mthd and to analyze the impact of its key design components, guided by the following research questions:

\begin{itemize}[leftmargin=1em]
    \item \textbf{RQ1 (overall performance)}: How does \mthd compare with strong baseline methods in predictive performance?
    \item \textbf{RQ2 (temporal modeling)}: How does temporal modeling affect performance and the learned multi-scale attention patterns?
    \item \textbf{RQ3 (spatial retrieval)}: How do monthly and yearly spatial retrievals influence prediction and help capture heterogeneity?
\end{itemize}

\subsection{Datasets}
\label{sec:datasets}

We evaluate \mthd on terrestrial ecosystem flux prediction for three key carbon-cycle targets: CO$_2$, GPP, and CH$_4$.
All datasets are organized into a unified site-year format, where each sample corresponds to one calendar year at a fixed location, with hierarchical environmental variables and a daily flux target.
Process-based simulations provide dense spatiotemporal coverage and enable learning across diverse environmental regimes.
We include agricultural carbon flux simulations  (CO$_2$, GPP) generated by the \textit{ecosys}
 model~\cite{ecosys}, and global wetland methane simulations (CH$_4$) generated by the TEM-MDM model~\cite{TEM}. 
These products span heterogeneous climates, land-cover types, and management conditions, serving as the primary source for learning cross-scale spatiotemporal patterns.
To evaluate robustness under real-world measurement constraints, we use tower-based eddy-covariance datasets derived from the FLUXNET network. Specifically, we rely on FLUXNET-based CO$_2$ and GPP products from AgroFlux~\cite{agroflux}, curated from the FLUXNET2015 dataset~\cite{fluxnet2015}, and CH$_4$ flux products from X-MethaneWet~\cite{xmethanewet}, based on the FLUXNET-CH$_4$ dataset~\cite{fluxnet_ch4}. Compared with simulations, observation-based datasets are spatially sparse and contain stronger site-specific noise and missingness, providing a realistic testbed for model generalization.
Across all datasets, we adopt a temporal extrapolation protocol: models are trained on earlier years and evaluated on subsequent years. In simulation experiments, we additionally hold out an intermediate year as an auxiliary pool used only for yearly-level retrieval; for fairness, non–retrieval baselines are allowed to train on this year, so \mthd in fact uses fewer labeled years. In observation experiments, we use the same temporal extrapolation in a two-stage transfer setting: the model is first pre-trained on simulations and then fine-tuned on FLUXNET observations, while still using the simulation auxiliary pool for yearly retrieval. Further details are provided in Appendix~\ref{sec:appendix_dataset}.

\begin{table*}[htbp]
\centering
\caption{Performance comparison of CH$_4$ prediction on TEM-MDM and FLUXNET datasets (mean $\std{\text{std}}$).}
\vspace{-0.1in}
\label{tab:ch4_table}
\setlength{\tabcolsep}{1.5pt}
\resizebox{\textwidth}{!}{
\begin{tabular}{l|cccccccccccccc|cc}
\toprule
\multirow{3}{*}{\textbf{Model}} & \multicolumn{14}{c|}{\textbf{TEM-MDM}} & \multicolumn{2}{c}{\textbf{FLUXNET}} \\
\cmidrule(lr){2-15} \cmidrule(lr){16-17}
 & \multicolumn{2}{c}{\textbf{Global}} & \multicolumn{2}{c}{\textbf{N. America}} & \multicolumn{2}{c}{\textbf{S. America}} & \multicolumn{2}{c}{\textbf{Europe}} & \multicolumn{2}{c}{\textbf{Africa}} & \multicolumn{2}{c}{\textbf{Asia}} & \multicolumn{2}{c|}{\textbf{Oceania}} & \multicolumn{2}{c}{\textbf{CH$_4$}} \\
\cmidrule(lr){2-3} \cmidrule(lr){4-5} \cmidrule(lr){6-7} \cmidrule(lr){8-9} \cmidrule(lr){10-11} \cmidrule(lr){12-13} \cmidrule(lr){14-15} \cmidrule(lr){16-17}
 & RMSE $\downarrow$ & $R^2$ $\uparrow$ & RMSE $\downarrow$ & $R^2$ $\uparrow$ & RMSE $\downarrow$ & $R^2$ $\uparrow$ & RMSE $\downarrow$ & $R^2$ $\uparrow$ & RMSE $\downarrow$ & $R^2$ $\uparrow$ & RMSE $\downarrow$ & $R^2$ $\uparrow$ & RMSE $\downarrow$ & $R^2$ $\uparrow$ & RMSE $\downarrow$ & $R^2$ $\uparrow$ \\
\midrule
LSTM & $3.06 \std{0.06}$ & $0.94 \std{.01}$ & $3.12 \std{0.05}$ & $0.93 \std{.01}$ & $2.03 \std{0.03}$ & $0.75 \std{.03}$ & $1.84 \std{0.07}$ & $0.96 \std{.00}$ & $1.41 \std{0.05}$ & $0.90 \std{.01}$ & $3.49 \std{0.05}$ & $0.97 \std{.00}$ & $8.09 \std{0.55}$ & $0.76 \std{.02}$ & $54.48 \std{1.20}$ & $0.44 \std{.02}$ \\
EA-LSTM & $5.91 \std{0.26}$ & $0.76 \std{.01}$ & $5.51 \std{0.07}$ & $0.80 \std{.02}$ & $3.15 \std{0.06}$ & $0.39 \std{.05}$ & $3.04 \std{0.20}$ & $0.90 \std{.01}$ & $2.53 \std{0.27}$ & $0.68 \std{.01}$ & $7.94 \std{0.68}$ & $0.82 \std{.02}$ & $15.14 \std{1.27}$ & $0.17 \std{.01}$ & $57.09 \std{0.32}$ & $0.39 \std{.01}$ \\
P\_sLSTM & $8.80 \std{0.48}$ & $0.39 \std{.10}$ & $7.16 \std{0.15}$ & $0.61 \std{.06}$ & $6.00 \std{0.30}$ & $-1.20 \std{.34}$ & $5.02 \std{0.50}$ & $0.69 \std{.03}$ & $5.00 \std{0.22}$ & $-0.57 \std{.19}$ & $13.11 \std{0.95}$ & $0.45 \std{.11}$ & $18.58 \std{2.32}$ & $-0.57 \std{.05}$ & $62.84 \std{0.78}$ & $0.27 \std{.02}$ \\ 
\midrule
TCN & $7.43 \std{0.14}$ & $0.59 \std{.03}$ & $7.19 \std{1.02}$ & $0.67 \std{.03}$ & $3.76 \std{0.23}$ & $0.05 \std{.08}$ & $4.64 \std{1.19}$ & $0.77 \std{.04}$ & $3.51 \std{0.23}$ & $0.28 \std{.04}$ & $11.10 \std{1.04}$ & $0.62 \std{.02}$ & $13.78 \std{3.46}$ & $-0.14 \std{.01}$ & $82.04 \std{8.45}$ & $-0.26 \std{.26}$ \\ 
TimesNet & $13.24 \std{0.37}$ & $-0.20 \std{.03}$ & $13.02 \std{0.09}$ & $-0.13 \std{.12}$ & $7.03 \std{0.40}$ & $-2.03 \std{.05}$ & $8.38 \std{0.44}$ & $0.26 \std{.06}$ & $6.48 \std{0.68}$ & $-1.13 \std{.23}$ & $20.30 \std{1.17}$ & $-0.15 \std{.08}$ & $22.78 \std{0.69}$ & $-0.89 \std{.24}$ & $88.44 \std{3.75}$ & $-0.45 \std{.12}$ \\ 
\midrule
Transformer & $3.58 \std{0.08}$ & $0.90 \std{.01}$ & $3.80 \std{0.33}$ & $0.91 \std{.01}$ & $2.01 \std{0.07}$ & $0.73 \std{.01}$ & $2.29 \std{0.36}$ & $0.94 \std{.00}$ & $1.67 \std{0.07}$ & $0.84 \std{.01}$ & $4.00 \std{0.45}$ & $0.95 \std{.00}$ & $9.06 \std{1.86}$ & $0.50 \std{.04}$ & $55.87 \std{6.28}$ & $0.41 \std{.13}$ \\
iTransformer & $7.82 \std{0.14}$ & $0.52 \std{.02}$ & $7.75 \std{0.36}$ & $0.54 \std{.02}$ & $4.26 \std{0.41}$ & $-0.09 \std{.05}$ & $5.09 \std{0.60}$ & $0.69 \std{.02}$ & $3.06 \std{0.06}$ & $0.41 \std{.02}$ & $10.46 \std{0.60}$ & $0.65 \std{.03}$ & $17.98 \std{1.59}$ & $-0.48 \std{.09}$ & $83.24 \std{7.28}$ & $-0.30 \std{.22}$ \\
Pyraformer & $6.43 \std{0.16}$ & $0.72 \std{.01}$ & $6.19 \std{0.29}$ & $0.74 \std{.02}$ & $3.16 \std{0.19}$ & $0.39 \std{.04}$ & $5.00 \std{0.45}$ & $0.74 \std{.03}$ & $2.81 \std{0.11}$ & $0.60 \std{.05}$ & $8.85 \std{0.29}$ & $0.78 \std{.02}$ & $13.36 \std{0.20}$ & $0.35 \std{.11}$ & $63.66 \std{5.03}$ & $0.24 \std{.12}$ \\
DUET & $11.31 \std{0.19}$ & $0.13 \std{.00}$ & $10.86 \std{0.69}$ & $0.22 \std{.06}$ & $6.58 \std{0.08}$ & $-1.68 \std{.25}$ & $7.47 \std{0.34}$ & $0.41 \std{.05}$ & $5.01 \std{0.46}$ & $-0.27 \std{.05}$ & $16.90 \std{0.63}$ & $0.20 \std{.04}$ & $21.12 \std{1.39}$ & $-0.61 \std{.12}$ & $100.88 \std{11.48}$ & $-0.91 \std{.42}$ \\
PatchTST & $12.47 \std{0.45}$ & $-0.06 \std{.04}$ & $11.70 \std{0.45}$ & $0.09 \std{.07}$ & $6.59 \std{0.10}$ & $-1.69 \std{.36}$ & $7.76 \std{0.52}$ & $0.37 \std{.05}$ & $6.38 \std{0.49}$ & $-1.07 \std{.21}$ & $19.32 \std{0.99}$ & $-0.04 \std{.04}$ & $22.15 \std{1.25}$ & $-0.77 \std{.13}$ & $69.14 \std{3.14}$ & $0.11 \std{.08}$ \\ 
\midrule
DLinear & $12.66 \std{0.39}$ & $-0.09 \std{.04}$ & $12.09 \std{0.29}$ & $0.02 \std{.16}$ & $9.19 \std{0.42}$ & $-4.20 \std{.18}$ & $8.40 \std{0.65}$ & $0.26 \std{.08}$ & $7.71 \std{0.61}$ & $-2.01 \std{.11}$ & $17.97 \std{0.64}$ & $0.10 \std{.02}$ & $23.24 \std{1.38}$ & $-0.95 \std{.13}$ & $85.58 \std{0.95}$ & $-0.36 \std{.03}$ \\
TSMixer & $3.72 \std{0.09}$ & $0.89 \std{.01}$ & $3.41 \std{0.15}$ & $0.91 \std{.02}$ & $2.35 \std{0.07}$ & $0.67 \std{.03}$ & $2.03 \std{0.25}$ & $0.95 \std{.00}$ & $1.89 \std{0.16}$ & $0.78 \std{.03}$ & $3.89 \std{0.41}$ & $0.95 \std{.01}$ & $11.01 \std{0.81}$ & $0.44 \std{.04}$ & $61.39 \std{6.89}$ & $0.29 \std{.16}$ \\
TimeMixer & $12.66 \std{0.14}$ & $-0.26 \std{.05}$ & $12.80 \std{0.33}$ & $-0.25 \std{.09}$ & $7.15 \std{0.24}$ & $-2.14 \std{.71}$ & $8.04 \std{0.83}$ & $0.22 \std{.10}$ & $6.21 \std{0.16}$ & $-1.42 \std{.21}$ & $19.26 \std{0.27}$ & $-0.18 \std{.02}$ & $20.83 \std{1.96}$ & $-0.99 \std{.12}$ & $83.92 \std{4.12}$ & $-0.31 \std{.13}$ \\ 
\midrule
\mthd & $\textbf{2.08} \std{0.22}$ & $\textbf{0.97} \std{.01}$ & $\textbf{2.11} \std{0.25}$ & $\textbf{0.97} \std{.01}$ & $\textbf{1.22} \std{0.19}$ & $\textbf{0.91} \std{.03}$ & $\textbf{1.24} \std{0.12}$ & $\textbf{0.98} \std{.00}$ & $\textbf{0.96} \std{0.11}$ & $\textbf{0.95} \std{.01}$ & $\textbf{2.55} \std{0.23}$ & $\textbf{0.98} \std{.00}$ & $\textbf{5.20} \std{0.64}$ & $\textbf{0.90} \std{.03}$ & $\textbf{51.97} \std{0.77}$ & $\textbf{0.49} \std{.01}$ \\
\bottomrule
\end{tabular}
}
\vspace{-0.1in}
\end{table*}

\subsection{Experimental Setup}
\textbf{Baselines.}
We compare \mthd against representative temporal baselines spanning multiple modeling paradigms. Recurrent neural networks (RNNs) include LSTM~\cite{lstm}, EA-LSTM~\cite{EA-LSTM}, and P\_sLSTM~\cite{pslstm}. Convolutional and frequency-aware methods include TCN~\cite{tcn} and TimesNet~\cite{timesnet}. Transformer-based approaches include the standard Transformer~\cite{transformer}, Pyraformer~\cite{pyraformer}, DUET~\cite{duet}, PatchTST~\cite{patchtst}, and iTransformer~\cite{itransformer}. We further include DLinear~\cite{dlinear} as a linear decomposition baseline, as well as TSMixer~\cite{tsmixer} and TimeMixer~\cite{timemixer} as recent MLP-based architectures. All baselines are implemented using publicly available implementations and evaluated under consistent experimental settings, detailed in Appendix~\ref{sec:appendix_baseline}. For all baselines, monthly and yearly covariates are simply replicated to the daily resolution and concatenated with daily drivers

\textbf{Evaluation metrics.}
We evaluate model performance using Root Mean Squared Error (RMSE) and the coefficient of determination ($R^2$). RMSE measures absolute prediction accuracy in the physical units of each flux variable. To account for strong spatial heterogeneity in flux magnitudes, we adopt a \emph{within-site} $R^2$ formulation in which the baseline mean in the denominator is computed separately for each site, yielding a more discriminative measure of temporal prediction.
The within-site $R^2$ is defined as
$$
R^2 = 1 - \frac{\sum_i \sum_t \left( y_{i,t} - \hat{y}_{i,t} \right)^2}{\sum_i \sum_t \left( y_{i,t} - \bar{y}_i \right)^2},
$$
where $y_{i,t}$ and $\hat{y}_{i,t}$ denote the observed and predicted fluxes at site $i$ and time $t$, respectively, and $\bar{y}_i= \frac{1}{T_i} \sum_t y_{i,t}$ is the temporal mean of observations at site $i$. 

\subsection{Overall Performance (RQ1)}

We evaluate all models on a sequence of carbon flux prediction tasks that become progressively harder, starting from relatively smooth, data-rich CO$_2$ and GPP and moving to the highly heterogeneous and data-scarce CH$_4$. This setup lets us test which inductive biases are actually useful for ecosystem flux prediction, and we find that the process-informed design of \mthd is robust and generalizable across these diverse tasks.

\subsubsection{GPP and CO$_2$ prediction on Ecosys and FLUXNET}

\begin{table}[t]
\centering
\caption{GPP and CO$_2$ performance on Ecosys and FLUXNET (mean $\std{\text{std}}$), with best performance highlighted in bold.}
\label{tab:gpp_co2_table}
\setlength{\tabcolsep}{1.5pt}
\resizebox{0.48\textwidth}{!}{%
\begin{tabular}{l|cccc|cccc}
\toprule
\multirow{3}{*}{\textbf{Model}}  & \multicolumn{4}{c|}{\textbf{Ecosys}} & \multicolumn{4}{c}{\textbf{FLUXNET}} \\
\cmidrule(lr){2-5} \cmidrule(lr){6-9}
 & \multicolumn{2}{c}{\textbf{GPP}} & \multicolumn{2}{c|}{\textbf{CO$_2$}} & \multicolumn{2}{c}{\textbf{GPP}} & \multicolumn{2}{c}{\textbf{CO$_2$}} \\
\cmidrule(lr){2-3} \cmidrule(lr){4-5} \cmidrule(lr){6-7} \cmidrule(lr){8-9}
 & RMSE $\downarrow$ & $R^2$ $\uparrow$ & RMSE $\downarrow$ & $R^2$ $\uparrow$ & RMSE $\downarrow$ & $R^2$ $\uparrow$ & RMSE $\downarrow$ & $R^2$ $\uparrow$ \\
\midrule
LSTM        & $2.87 \std{.19}$ & $0.84 \std{.02}$ & $0.48 \std{.04}$ & $0.89 \std{.02}$ & $1.50 \std{.05}$ & $0.84 \std{.01}$ & $3.67 \std{.10}$ & $0.71 \std{.02}$ \\
EA-LSTM      & $3.48 \std{.21}$ & $0.77 \std{.03}$ & $0.65 \std{.02}$ & $0.83 \std{.01}$ & $2.26 \std{.28}$ & $0.64 \std{.09}$ & $4.46 \std{.24}$ & $0.56 \std{.05}$ \\ 
P\_sLSTM    & $3.85 \std{.84}$ & $0.70 \std{.13}$ & $0.69 \std{.13}$ & $0.87 \std{.18}$ & $1.94 \std{.07}$ & $0.74 \std{.02}$ & $4.06 \std{.16}$ & $0.64 \std{.03}$ \\ \midrule
TCN         & $3.35 \std{.27}$ & $0.79 \std{.03}$ & $0.60 \std{.01}$ & $0.86 \std{.00}$ & $1.52 \std{.07}$ & $0.84 \std{.02}$ & $3.69 \std{.20}$ & $0.70 \std{.03}$ \\
TimesNet    & $3.73 \std{.18}$ & $0.73 \std{.02}$ & $0.92 \std{.08}$ & $0.65 \std{.06}$ & $1.75 \std{.11}$ & $0.79 \std{.03}$ & $3.94 \std{.13}$ & $0.66 \std{.02}$ \\ \midrule
Transformer & $4.76 \std{.11}$ & $0.57 \std{.02}$ & $0.60 \std{.03}$ & $0.85 \std{.01}$ & $2.25 \std{.03}$ & $0.65 \std{.01}$ & $4.75 \std{.03}$ & $0.51 \std{.01}$ \\
iTransformer & $4.15 \std{.59}$ & $0.67 \std{.08}$ & $0.90 \std{.04}$ & $0.64 \std{.08}$ & $1.87 \std{.01}$ & $0.75 \std{.01}$ & $4.26 \std{.01}$ & $0.60 \std{.01}$ \\
Pyraformer  & $3.16 \std{.18}$ & $0.81 \std{.02}$ & $0.61 \std{.02}$ & $0.85 \std{.01}$ & $1.52 \std{.06}$ & $0.84 \std{.01}$ & $3.45 \std{.02}$ & $0.74 \std{.00}$ \\
DUET        & $4.22 \std{.53}$ & $0.66 \std{.08}$ & $0.88 \std{.18}$ & $0.68 \std{.13}$ & $1.49 \std{.03}$ & $0.85 \std{.01}$ & $3.81 \std{.15}$ & $0.68 \std{.02}$ \\
PatchTST    & $5.61 \std{.22}$ & $0.40 \std{.05}$ & $1.10 \std{.25}$ & $0.50 \std{.23}$ & $1.61 \std{.10}$ & $0.82 \std{.02}$ & $4.29 \std{.24}$ & $0.60 \std{.05}$ \\ \midrule
DLinear     & $4.58 \std{.58}$ & $0.59 \std{.10}$ & $1.16 \std{.01}$ & $0.59 \std{.10}$ & $1.82 \std{.19}$ & $0.77 \std{.05}$ & $3.91 \std{.18}$ & $0.67 \std{.03}$ \\ 
TSMixer     & $3.77 \std{.51}$ & $0.72 \std{.07}$ & $0.75 \std{.10}$ & $0.76 \std{.06}$ & $1.50 \std{.03}$ & $0.84 \std{.01}$ & $3.56 \std{.06}$ & $0.72 \std{.01}$ \\
TimeMixer   & $3.77 \std{.07}$ & $0.73 \std{.01}$ & $0.84 \std{.06}$ & $0.71 \std{.04}$ & $1.56 \std{.01}$ & $0.83 \std{.00}$ & $3.69 \std{.07}$ & $0.70 \std{.01}$ \\ \midrule
\mthd       & $\textbf{2.60} \std{.11}$ & $\textbf{0.87} \std{.01}$ & $\textbf{0.36} \std{.02}$ & $\textbf{0.95} \std{.02}$ & $\textbf{1.39} \std{.04}$ & $\textbf{0.87} \std{.02}$ & $\textbf{3.15} \std{.05}$ & $\textbf{0.78} \std{.01}$ \\
\bottomrule
\end{tabular}%
}
\end{table}

We first evaluate the overall performance of \mthd on GPP and CO$_2$ prediction using the AgroFlux benchmark, which includes both simulation-based (Ecosys) and observation-based (FLUXNET) datasets. These fluxes are generally regarded as comparatively easier targets due to their smoother temporal dynamics and higher signal-to-noise ratios, and thus provide a natural starting point for assessing model behavior.
As summarized in Table~\ref{tab:gpp_co2_table}, \mthd consistently achieves the lowest RMSE and highest $R^2$ across both datasets and fluxes. Among baselines, RNN-based models perform strongly, reinforcing the view that carbon flux prediction is essentially a state-response problem that benefits from time-step-wise conditioning on instantaneous drivers. In contrast, CNN-style models that emphasize complex multi-periodic patterns do not outperform simple RNNs, suggesting that such inductive biases are less crucial in these externally forced systems.
Models that enforce channel independence (e.g., PatchTST, DLinear) perform markedly worse, indicating that ecosystem fluxes depend on multivariate interactions that are lost when covariates are modeled separately. DUET partly addresses multivariate structure, but its grouping variables by statistical similarity is misaligned with physically meaningful roles, so functionally complementary drivers (e.g., precipitation vs. radiation) may be sparsified as noise. iTransformer preserves cross-variable correlations but compresses temporal information into global representations, weakening fine-grained alignment between meteorological drivers and fluxes and leading to suboptimal modeling of transient events compared to time-step-aware architectures.

\begin{figure*}[htbp]
    \centering
    \begin{subfigure}{0.4\textwidth}
        \centering
        \includegraphics[width=\linewidth]{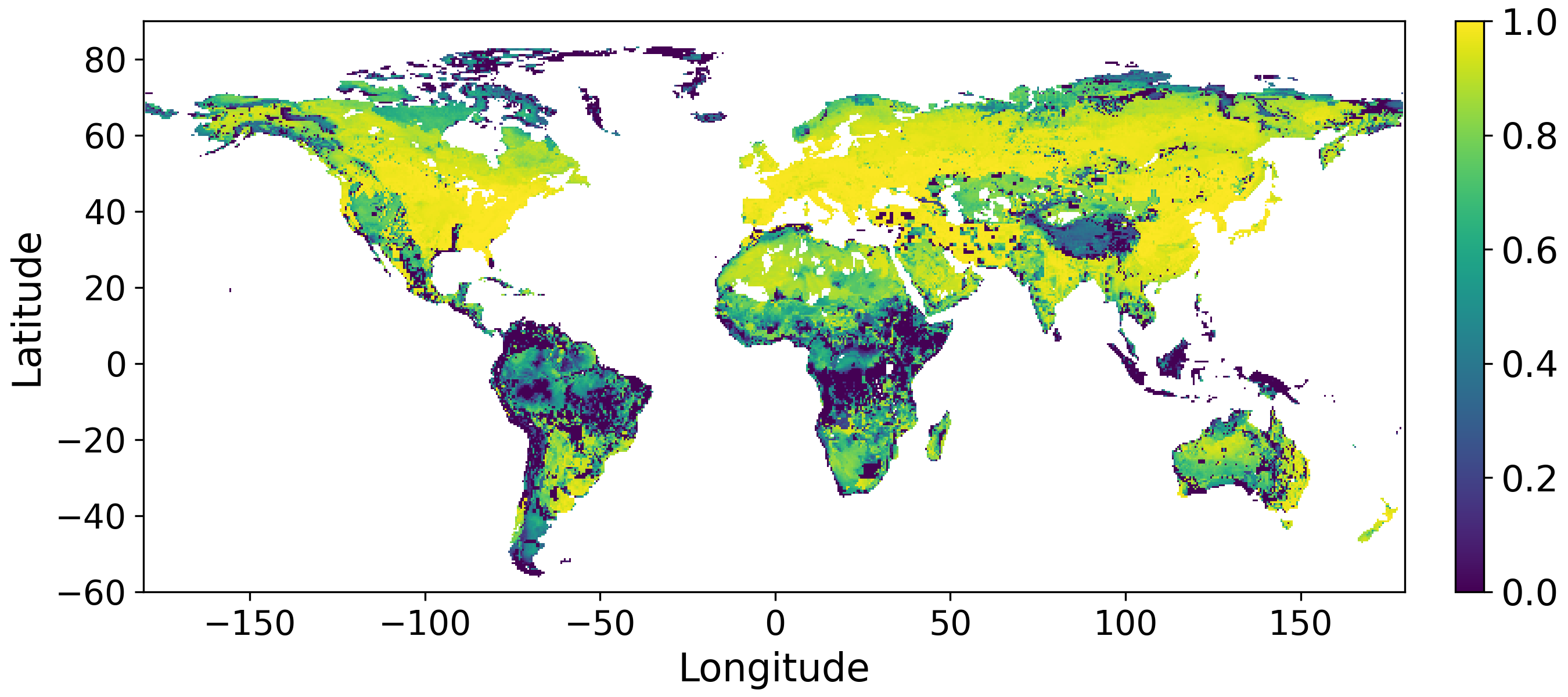}
        \label{fig:FLUXNET_temp_ts}
    \end{subfigure}
    \begin{subfigure}{0.4\textwidth}
        \centering
        \includegraphics[width=\linewidth]{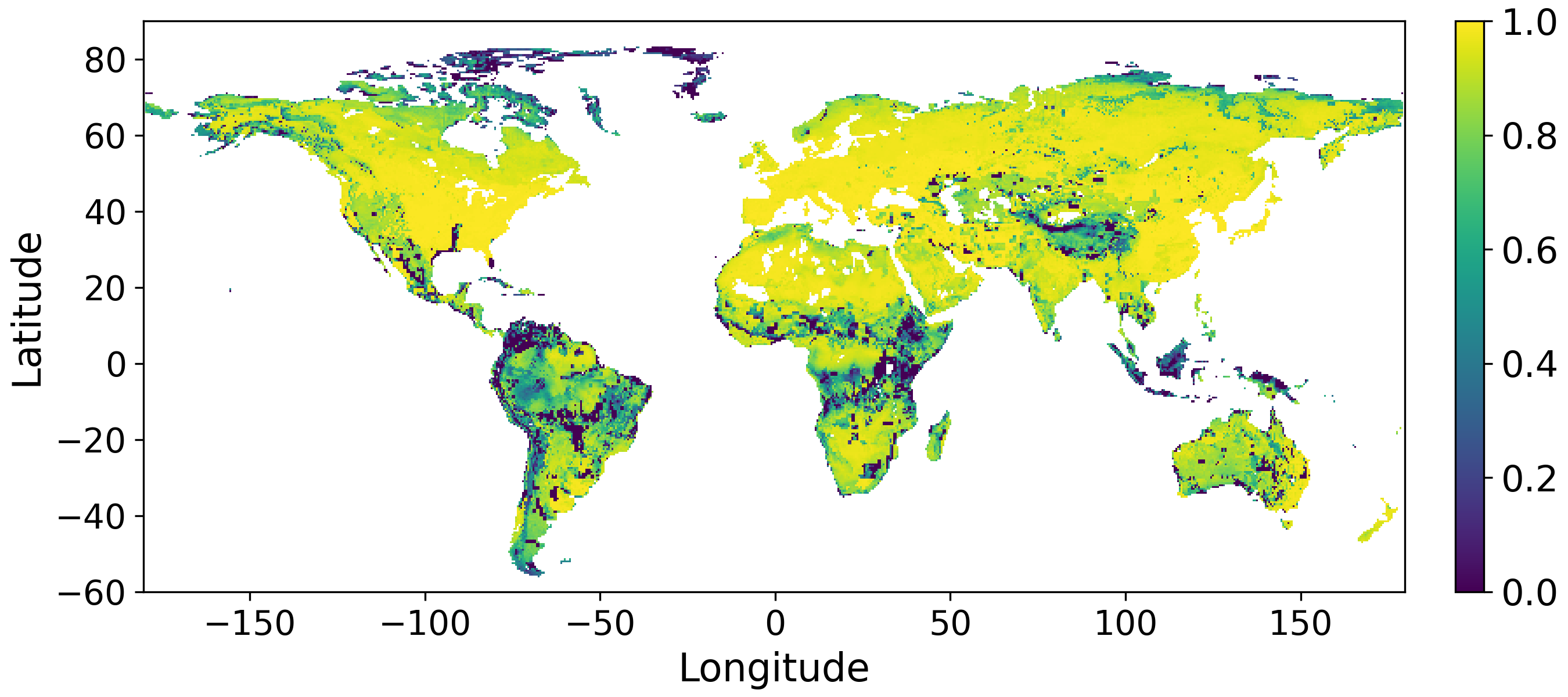}
        \label{fig:FLUXNET_spatial_ts}
    \end{subfigure}
    \vspace{-0.25in}
    \caption{Spatial distribution of point-wise $R^2$ on TEM-MDM in test year 2018, comparing LSTM (left) and \mthd (right).}
    \vspace{-0.1in}
    \label{fig:TEM_r2_comparison}
\end{figure*}

\begin{figure}[htbp]
\centering
\includegraphics[width=\linewidth]{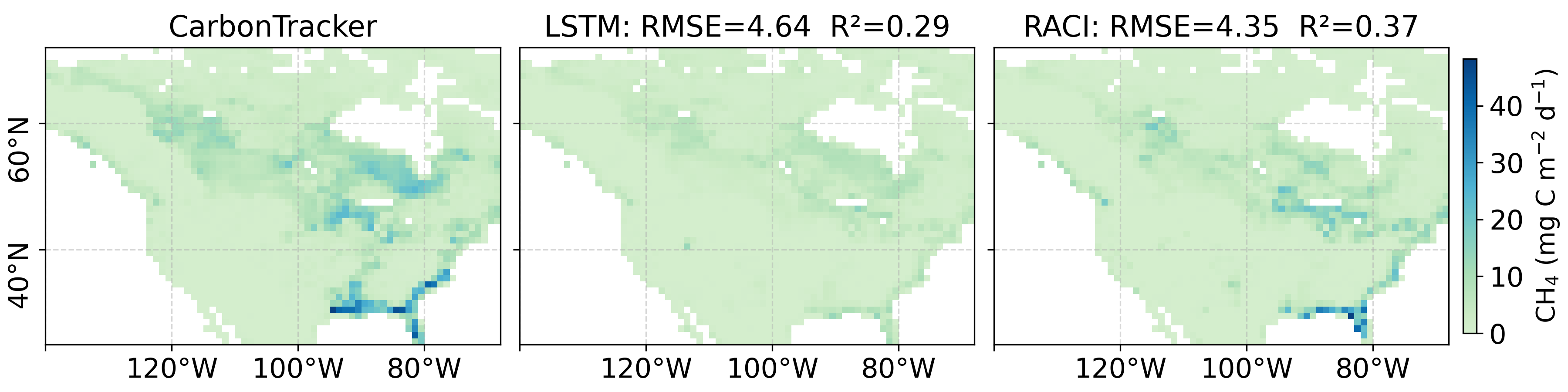}
\vspace{-0.15in}
\caption{Annual mean CH$_4$ emissions over North America in 2018 from CarbonTracker, LSTM, and \mthd prediction.}
\vspace{-0.15in}
\label{fig:carbontracker_mean}
\end{figure}

\subsubsection{$CH_4$ prediction on TEM-MDM and FLUXNET-$CH_4$}

Unlike CO$_2$ and GPP, global CH$_4$ flux modeling in TEM-MDM is substantially more challenging due to the nonlinear biophysical processes governing methane production and oxidation. At the global scale, the predictor must capture strong spatial heterogeneity across climatic zones and wetland types, making this task a stringent test of whether architectures can learn complex process-level interactions from dense but diverse simulation data.
The global CH$_4$ results in Table~\ref{tab:ch4_table} show that models that already appeared fragile for GPP and CO$_2$ often obtain negative $R^2$ scores, and even architectures that performed reasonably well on those fluxes degrade substantially. Among the stronger baselines, contrasts between close variants further pinpoint problematic assumptions: TimeMixer underperforms TSMixer, suggesting that explicit trend–seasonality decomposition can blur the short, threshold-triggered “hot moments’’ that dominate CH$_4$ emissions; P-sLSTM falls behind standard LSTM, indicating that patching may interrupt the continuous build-up of anaerobic conditions; and the decline of Pyraformer suggests that pyramidal coarsening tends to smooth out sharp emission spikes. Overall, these patterns indicate that for complex fluxes such as CH$_4$, architectures with rigid structural assumptions are less suitable than simpler continuous-state models like LSTM and \mthd. Beyond global averages, performance also varies strongly across continents: even LSTM performs poorly in South America, Africa, and Oceania, as shown numerically in Table~\ref{tab:ch4_table} and spatially in the point-wise $R^2$ map in Figure~\ref{fig:TEM_r2_comparison}. This spatial imbalance is largely mitigated by \mthd, which maintains balance across regions.

The challenge becomes even more severe for real-world CH$_4$ observations. The X-MethaneWet dataset contains only 30 sites globally, with strong spatial bias: towers are concentrated in North America and Europe, while much of the Southern Hemisphere remains unmonitored. Under such extreme data sparsity and geographic imbalance, many ML models experience representation collapse when trained directly on these observations, as shown in Table \ref{tab:ch4_table}. We observe a clear dichotomy: several recent architectures designed for general time-series forecasting may yield negative $R^2$ scores on FLUXNET-CH$_4$ data, whereas simpler baselines such as standard LSTM, Transformer, and the domain-informed EA-LSTM maintain reasonable performance. This contrast highlights the role of inductive bias: sophisticated models rely on structural assumptions that are well-suited to dense, homogeneous datasets but become mismatched for sparse, highly heterogeneous methane fluxes. Vanilla architectures, by imposing weaker structural constraints, are less prone to such failures and can still capture basic sequential dependencies. Building on this baseline robustness, \mthd introduces a domain-aligned inductive bias by leveraging role-aware retrieval to transfer information from regime-similar simulation patterns, thereby bridging observational gaps and achieving the strongest generalization among all evaluated models. Representative time series comparisons in Figure~\ref{fig:FLUXNET_ts} in Appendix~\ref{sec:appendix_fluxnet_ts} further illustrate these behaviors at individual sites.

\subsubsection{Out-of-distribution evaluation on CarbonTracker}

Finally, we conduct an out-of-distribution evaluation by comparing our upscaled predictions against CarbonTracker~\cite{carbontracker}, a high-fidelity top-down atmospheric inversion product often used as a proxy for regional-to-global flux dynamics. Because FLUXNET-CH$4$ towers are extremely sparse and geographically imbalanced, with most sites located in North America, which is itself biophysically diverse, we restrict our out-of-distribution evaluation to this region. We adopt a two-stage protocol: the model is first pre-trained on TEM-MDM simulations, then fine-tuned on FLUXNET sites in North America, and finally evaluated on the North American CarbonTracker grid, directly testing whether the framework can generalize from a small set of regional tower measurements to a continuous regional flux field. This comparison between bottom-up inference and top-down atmospheric constraints provides a stringent assessment of RACI’s scalability and its potential for large-scale carbon cycle monitoring. 
Figure~\ref{fig:carbontracker_mean} illustrates a key limitation of the baseline LSTM: its predictions appear overly smooth in regions without towers and blurring critical emission hotspots (e.g., Southeast U.S. wetlands). In contrast, \mthd better reconstructs the spatial heterogeneity of methane fluxes, better localizes high-emission zones and achieves a higher $R^2$, indicating that its role-aware retrieval provides spatial priors that improve extrapolation from sparse towers.

\subsection{Impact of Temporal Modeling (RQ2)}
\subsubsection{Ablation on temporal modeling}
To assess the effectiveness of Role-Separating Temporal Modeling, we construct a variant (-Temporal) that disables the hierarchical temporal encoder while keeping both spatial retrieval modules unchanged. In this variant, coarse embeddings are obtained by simple temporal averaging, and coarse-scale information is propagated to finer scales via naive value replication, without learned encoder–decoder interactions. We evaluate this ablation on the TEM-MDM dataset, which is both demanding and sufficiently large to support robust comparison of variants. As shown in Table~\ref{tab:ablation}, the -Temporal variant exhibits a clear degradation in performance, indicating that retrieval alone is not sufficient: it relies on informative, role-aware temporal embeddings. These results support our encoder–decoder design as a key inductive bias that organizes the latent space into scale-consistent representations that can be meaningfully compared and retrieved.

\subsubsection{Interpreting multi-scale temporal attention in \mthd}
\label{sec:temporal_attention_main}
The ablation results above show that temporal embeddings are not equivalent to simple averaging and replication. To understand how \mthd organizes information across scales, we examine attention weights in the aggregation and propagation blocks for CH$_4$ after pre-training on TEM-MDM and fine-tuning on FLUXNET-CH$_4$, so that the patterns reflect a realistic setting. In aggregation, attention selects which fine-scale steps are preserved when forming coarse summaries; in propagation, it controls how regime-level context is injected back into finer resolutions. Because these weights are computed via cross-attention on input features, their patterns provide a proxy for feature importance at each timescale. Representative examples are shown in Figure~\ref{fig:temporal_attention} in Appendix~\ref{sec:appendix_temporal_attention}.

For monthly-to-yearly aggregation, Northern Hemisphere sites systematically receive higher attention on winter months. This indicates that yearly regime descriptors emphasize seasonal contrast: warm seasons often exhibit uniformly high methane emissions, whereas colder periods show more distinct behavior and are therefore more informative for separating regimes. For daily-to-monthly aggregation, we correlate attention weights with key drivers (precipitation, solar radiation, air temperature, and vapor pressure). Rather than distributing attention evenly, the model tends to focus on whichever variable constrains local flux, consistent with a “law of the minimum’’ behavior. For example, at FI-Lom (Finland), a subarctic peatland with only 484~mm annual rainfall, attention peaks on precipitation ($r=0.70$), reflecting sensitivity to episodic water input; at US-LA1 (Louisiana), a humid subtropical marsh with 1625~mm rainfall and tidal influence, attention shifts to temperature ($r=0.53$), indicating energy-limited microbial activity. In contrast, solar radiation rarely dominates: its fixed seasonality provides limited additional information and is seldom the limiting factor during peak seasons. Overall, the model emphasizes drivers with the greatest ecological leverage under local constraints, rather than any single variable globally.

In the coarse-to-fine propagation modules, we observe a complementary pattern. Both yearly-to-monthly and monthly-to-daily propagation assign larger gates to time steps with higher emission potential--typically warm, wet periods--while down-weighting less active periods. Rather than treating conditioners as a constant bias, \mthd uses them as adaptive priors that selectively modulate dynamic drivers when methane-favorable conditions are met. These attention patterns are consistent with the intended design: regime-level context shapes local dynamics most strongly when and where it matters for CH$_4$ emissions.

\begin{table}[t]
\centering
\caption{Ablation study of \mthd variants on TEM-MDM.}
\label{tab:ablation}
\resizebox{0.37\textwidth}{!}{
\begin{tabular}{lccccc}
\toprule
{Metric} & {Full} & {-Temporal} & {-Monthly} & {-Yearly} & {-Both} \\
\midrule
RMSE            & 2.08  & 3.08  & 2.28  & 3.19  & 3.31 \\
R$^2$           & 0.97  & 0.94  & 0.96  & 0.93  & 0.92 \\
\bottomrule
\end{tabular}
}
\vspace{-0.2in}
\end{table}

\subsection{Effect of Role-Aware Spatial Retrieval (RQ3)}
\subsubsection{Ablation on monthly and yearly retrieval}
We first assess the contribution of spatial retrieval at different scales. We construct three variants, {-Monthly}, {-Yearly} and {-Both}, by disabling the corresponding retrieval module while keeping the rest of the architecture unchanged. As shown in Table~\ref{tab:ablation}, removing either module degrades performance, with a markedly larger drop for {-Yearly}. Without yearly retrieval, the model still observes dynamic drivers but no longer accesses conditioner information that governs how these drivers translate into flux responses, and its predictions revert toward a global average behavior. In contrast, removing monthly retrieval weakens the ability to contextualize local driver signals with regional seasonal patterns, leading to smaller yet consistent performance losses. These results support the complementary roles of monthly (driver-oriented) and yearly (conditioner-oriented) retrieval in handling forcing and response shifts.

\subsubsection{Functional rather than geographic retrieval}
\label{sec:yearly_retrieval_main}
We next examine whether yearly-level retrieval is driven by ecological function rather than geographic proximity. Figure~\ref{fig:la_retrieval} in Appendix~\ref{sec:appendix_spatial_retrieval} shows a representative example in the Mississippi River Delta, where US-LA1 and US-LA2 are geographically adjacent and share similar climate (humid subtropical, $\sim$20$^\circ$C, $\sim$1600~mm/year), yet induce distinct retrieval locations. US-LA1, a brackish marsh with natural tidal dynamics, primarily retrieves subtropical wetland systems in South America. By contrast, US-LA2, a freshwater marsh influenced by the Davis Pond diversion project, preferentially retrieves managed agroecosystems in regions such as Saudi Arabia and Australia. This contrast indicates that \mthd uses yearly embeddings to distinguish natural versus engineered hydrological regimes and conditions predictions on inferred ecological roles, rather than relying solely on spatial distance or coarse climate similarity.

\section{Related Work}

\textbf{Process-based modeling.} Process-based models (PBMs) have long been the standard for terrestrial carbon and methane flux modeling, formulating ecosystem exchange through process-level equations grounded in physical, chemical, and biological principles. Land surface and ecosystem models such as CLM~\cite{CLM}, ORCHIDEE~\cite{ORCHIDEE}, Biome-BGC~\cite{running1993generalization}, LPJ~\cite{sitch2003evaluation}, and DAYCENT~\cite{grosso2000general,del2005daycent} explicitly simulate photosynthesis, respiration, and carbon–water–energy coupling to estimate GPP, NEE, and related carbon fluxes. For methane emissions, wetland PBMs including the Walter–Heimann model~\cite{walter2000process}, TEM-MDM~\cite{TEM}, DLEM~\cite{DLEM,DLEM2} and ELM-ECA~\cite{riley2011barriers} further resolve methane production, oxidation, and transport via diffusion, ebullition, and plant-mediated pathways. While these models offer strong physical interpretability and have been widely applied across ecosystems and climates, they depend on rigid structural assumptions and semi-empirical parameterizations tuned to specific environments, which limits their transferability under global heterogeneity. Their computational cost and fixed temporal structures also make it difficult to capture short-term variability, regime shifts, and scale-dependent ecosystem responses.

\textbf{Data-driven and hybrid models.}
Machine learning has substantially advanced terrestrial flux prediction, especially for upscaling eddy-covariance measurements to larger scales~\cite{tramontana2016predicting,jung2020scaling}. Deep models (CNNs, RNNs, Transformers) have been used to predict carbon fluxes from meteorological and remote-sensing inputs~\cite{dou2018estimating,dou2018comprehensive,reichstein2019deep,besnard2019memory,nathaniel2023metaflux,liu2024knowledge,cheng2025knowledge,chen2024quantifying}, often outperforming PBMs where observations are dense, but degrading under covariate shift and sparse tower coverage~\cite{reichstein2019deep,guevara2021machine}. Hybrid physics–ML frameworks alleviate part of this by enforcing physical constraints or regularizing with process-based simulations~\cite{tramontana2020partitioning,baghirov2025h2cm,rong2025hybrid}. However, most approaches still assume a single global mapping from drivers to fluxes, which under strong ecosystem heterogeneity causes predictions to regress toward biased global averages and perform poorly in unmonitored regimes.

\section{Conclusion}

In this paper, we introduced a Role-Aware Conditional Inference (\mthd) framework for terrestrial carbon and methane flux prediction. \mthd separates slow background conditioners from fast-varying drivers and employs role-aware spatial retrieval so that predictions are conditioned on both regional forcing patterns and regime-similar ecosystems, rather than relying on a single global response. Empirical results across simulation and observation benchmarks show consistent gains over a broad set of baseline models. As a general framework for role-aware retrieval-augmented spatiotemporal modeling, \mthd has potential beyond the specific fluxes and datasets studied here, offering a path toward more robust and interpretable AI for Earth system monitoring.



\bibliographystyle{ACM-Reference-Format}
\bibliography{ref}

\appendix
\newpage
\section{Process Background and Role Definitions}
\label{sec:process_appendix}

Process-based ecosystem models compute carbon and methane fluxes as explicit functions of environmental forcing, internal state variables, and slowly varying site properties. In many land surface and ecosystem models (e.g., CLM, Biome-BGC, LPJ), fluxes are written as products or combinations of response functions to different factors, such as solar radiation, temperature, soil moisture, vegetation state, and soil conditions. This structure implicitly distinguishes between fast-varying drivers, which respond on hourly to seasonal time scales to meteorological forcing, and slow background conditioners, which evolve over years to decades and set the long-term capacity of the ecosystem to exchange carbon with the atmosphere. 

For example, gross primary productivity (GPP) are often expressed as a light-use efficiency term multiplied by limiting functions of air temperature, vapor pressure deficit, and soil moisture, all of which can fluctuate strongly from day to day. At the same time, this instantaneous uptake is conditioned on slowly varying vegetation properties such as leaf area index, canopy nitrogen content, and plant functional type, which control the maximum attainable photosynthetic capacity but change only over phenological or successional time scales. Similarly, net ecosystem exchange (NEE/CO$_2$) depend on fast-varying soil temperature and moisture as drivers, while being conditioned on the size and turnover of soil carbon pools, soil texture, and nutrient status. In both cases, the flux can be viewed as the response of a relatively slowly evolving background state (carbon pools, vegetation and soil properties) to fast meteorological forcing.

This role separation is made explicit in the methane dynamics module of the Terrestrial Ecosystem Model (TEM-MDM)~\cite{TEM}. There, the gross methane production rate $M_P(z,t)$ at soil depth $z$ and time $t$ is written as a product of response functions to several controlling factors:
$$
M_{P}(z,t) = f(M_{ST}(z,t)) \cdot f(R_{X}(z,t)) \cdot f(S_{OM}(z,t)) \cdot M_{G0} \cdot f(pH).
$$

Here $M_{ST}(z,t)$ denotes the soil temperature and moisture state, and $R_{X}(z,t)$ denotes the redox status of the soil column (driven by the position of the water table and oxygen availability). Both are updated on (sub-)daily time scales in response to meteorological forcing and hydrological fluctuations. Soil organic matter $S_{OM}(z,t)$ is a somewhat more slowly evolving state variable, updated at monthly time scales from net primary productivity: it still varies in time, but integrates past productivity and therefore evolves more smoothly than the instantaneous meteorological drivers. In contrast, $M_{G0}$ is an ecosystem-specific baseline methanogenic potential and soil pH is a static, site-specific property. Together, $M_{G0}$ and $pH$ set the long-term capacity of the system to produce methane, while $M_{ST}(z,t)$, $R_X(z,t)$, and $S_{OM}(z,t)$ control the realized production on daily-to-monthly time scales around this capacity.

This decomposition naturally aligns with our terminology of \emph{drivers} and \emph{conditioners}: $M_{ST}(z,t)$, $R_{X}(z,t)$, and $S_{OM}(z,t)$ act as dynamic drivers that determine how methane production responds to short-term meteorological, hydrological, and substrate-availability changes, whereas $M_{G0}$ and $pH$ act as background conditioners that determine whether a site can sustain high emissions at all and how large the potential response can be.

\section{Datasets}
\label{sec:appendix_dataset}

This section provides detailed descriptions of the datasets, variables, and preprocessing procedures used in this study.
All datasets are harmonized into a unified site-year representation, consistent with the problem formulation in the main paper.

\subsection{AgroFlux CO$_2$ and GPP Datasets}
\label{sec:appendix_agroflux}

We use the CO$_2$ and GPP components of the AgroFlux benchmark~\cite{agroflux}, which combines process-based simulations from Ecosys~\cite{ecosys} with FLUXNET~\cite{fluxnet2015} eddy-covariance observations.

\paragraph{Simulation component (Ecosys)}
The Ecosys dataset provides daily outputs from 2000--2018 for 99 counties across Iowa, Illinois, and Indiana in the U.S.\ Corn Belt, under multiple management scenarios. Each county is simulated under 20 nitrogen fertilization rates ranging from 0 to 33.6 g N m$^{-2}$, representing diverse management intensities within otherwise identical soil and climate conditions. In our role terminology, these fertilization scenarios define slow-varying \emph{regime conditioners} that shape background carbon cycling states (e.g., soil organic carbon and plant productivity), while daily weather drives fast variability.

\paragraph{Observation component (FLUXNET)}
The AgroFlux CO$_2$/GPP observation dataset curated from FLUXNET2015~\cite{fluxnet2015} consists of 11 cropland eddy-covariance sites (US-Bo1, US-Bo2, US-Br1, US-Br3, US-IB1, US-KL1, US-Ne1, US-Ne2, US-Ne3, US-Ro1, US-Ro5) distributed across Illinois, Iowa, Michigan, Nebraska, and Minnesota, with site records spanning 2000–2020 at daily resolution.

\paragraph{Variables and role mapping}
We adopt the AgroFlux driver set as inputs and use daily GPP and CO$_2$ flux as prediction targets. Meteorological variables (e.g., precipitation, air temperature, radiation, vapor pressure deficit) are treated as fast \emph{drivers}. Static or slowly varying variables—soil bulk density (TBKDS), sand/silt content (TCSAND, TCSILT), soil pH (TPH), soil organic carbon (TSOC), crop type (PLANTT), and fertilization rate (FERTZR\_N)—are treated as \emph{conditioners} that define the local regime. This mapping is applied consistently across both Ecosys and FLUXNET-based datasets.

\paragraph{Preprocessing and temporal splits}
Following AgroFlux, all time series are segmented into yearly sequences of length 365 and normalized feature-wise. Missing observations in FLUXNET-based data are masked out in the loss and metrics, so models are evaluated only on available tower measurements. For Ecosys, we use a temporal extrapolation setup with 2000--2014 for training, 2015 reserved as a held-out auxiliary pool for retrieval, and 2016--2018 for testing. For the FLUXNET-based CO$_2$/GPP data, we follow the AgroFlux temporal split, training on 2000--2015 and testing on 2016--2020, while using the same 2015 Ecosys auxiliary pool for yearly-level retrieval. All experiments in the main text that involve AgroFlux use exactly these splits.

\begin{figure*}[htbp]
\centering
\includegraphics[width=\linewidth]{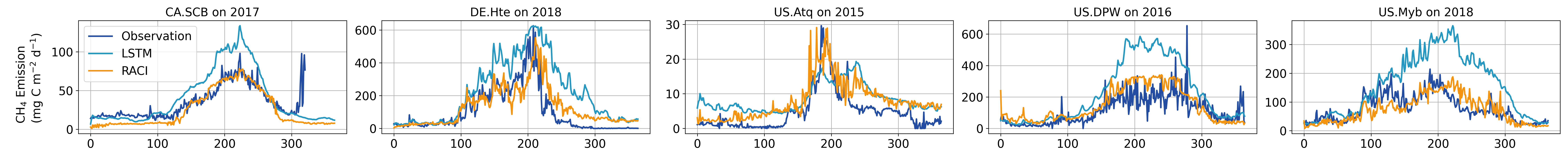}
\caption{Comparison of daily CH$_4$ emission time series from observations and from LSTM and \mthd predictions at five FLUXNET-CH$_4$ site-year sequences in the test set.}
\vspace{-0.15in}
\label{fig:FLUXNET_ts}
\end{figure*}

\subsection{X-MethaneWet CH$_4$ Benchmark}
\label{sec:appendix_xmethanewet}

For methane, we follow the X-MethaneWet benchmark~\cite{xmethanewet}, which couples process-based simulations from TEM-MDM~\cite{TEM} with tower observations from FLUXNET-CH$_4$~\cite{fluxnet_ch4}.

\paragraph{Simulation component (TEM-MDM)}
The TEM-MDM component provides daily CH$_4$ emissions and environmental variables at 0.5$^\circ$ global resolution from 1979--2018 for 62{,}470 wetland grid cells. Input features include four daily meteorological drivers (PREC, TAIR, SOLR, VAPR from ERA-Interim), a monthly net primary productivity (NPP) series, yearly global CO$_2$ and CH$_4$ concentrations, and static wetland/land-surface properties (wetland type, vegetation and plant functional type, soil texture fractions, soil bulk density, pH, elevation, and climate class).

\paragraph{Observation component (FLUXNET-CH$_4$)}
The FLUXNET-CH$_4$ component consists of 30 vegetated wetland sites (bog, fen, marsh, swamp, wet tundra, drained, and salt marsh) spanning Arctic–boreal, temperate, and (sub)tropical climate zones, with daily CH$_4$ fluxes typically between 2006 and 2018. For each site we retain the standardized FLUXNET-CH$_4$ meteorological drivers (air temperature, precipitation, vapor pressure) and align the tower location to the nearest TEM-MDM wetland grid cell to complete the feature set and wetland-type labels, following the construction in X-MethaneWet.

\paragraph{Variables and role mapping}
All TEM-MDM and FLUXNET-CH$_4$ records are converted into site-year sequences of length 365. In our framework we treat fast-varying meteorology and NPP as \emph{drivers}, and slowly varying soil, vegetation, climate, wetland type, and global gas concentrations as \emph{conditioners}.

\paragraph{Preprocessing and temporal splits}
Missing daily CH$_4$ fluxes in FLUXNET-CH$_4$ are masked in the loss and evaluation metrics, so models are only scored where tower data are available. For TEM-MDM, baseline models are trained on 1979--2008 and tested on 2009--2018. For \mthd, we train on 1979--2007, reserve 2008 as a held-out auxiliary pool used only for yearly-level retrieval, and still test on 2009--2018; thus baselines see one additional labeled year. For the FLUXNET-CH$_4$ data, we use the last available year of each site for testing and all previous years for training, while using the same 2008 TEM-MDM auxiliary pool for yearly-level retrieval. All experiments in the main text that involve X-MethaneWet use exactly these splits.

\section{Supplementary of Experiments}

\subsection{Baselines}
\label{sec:appendix_baseline}
In this section, we provide a brief overview of the baseline methods for reference in our paper.

\begin{itemize}[leftmargin=1em]
    \item LSTM~\cite{lstm}: A canonical recurrent neural network that models temporal dependencies via gated cell states, serving as a strong point-wise sequential baseline.
    \item EA-LSTM~\cite{EA-LSTM}: An LSTM variant originally proposed for hydrological modeling that injects static catchment attributes into the input gate, so that static properties (e.g., soil type, topography) modulate the recurrent dynamics.
    \item P\_sLSTM~\cite{pslstm}: A recent LSTM variant that applies temporal patching and modified gating to improve long-horizon forecasting and enable more parallel training over long sequences.
    \item TCN~\cite{tcn}: A temporal convolutional network that uses causal convolutions to obtain a large receptive field and capture long-range temporal dependencies.
    \item TimesNet~\cite{timesnet}: A general-purpose model that maps 1D time series into 2D tensors based on learned multi-periodicity and then applies standard 2D convolutional blocks.
    \item Transformer~\cite{transformer}: The vanilla Transformer architecture, using self-attention to model global temporal dependencies across all time steps.
    \item iTransformer~\cite{itransformer}: An inverted Transformer that first embeds each variable’s full time series into a single token and then applies attention over variables, explicitly emphasizing multivariate correlations.
    \item Pyraformer~\cite{pyraformer}: A Transformer variant with pyramidal attention that aggregates information in a multi-scale manner while reducing complexity on long sequences.
    \item DUET~\cite{duet}: An architecture that performs dual clustering and sparsification to selectively connect variables and time steps, aiming to model complex multivariate dependencies while filtering noise.
    \item PatchTST~\cite{patchtst}: A Transformer-based model that segments time series into patches and processes each channel independently, targeting long-term semantic dependencies with a channel-wise inductive bias.
    \item DLinear~\cite{dlinear}: A simple linear baseline that decomposes each univariate series into trend and seasonal components and fits independent linear mappings for each channel.
    \item TSMixer~\cite{tsmixer}: An all-MLP architecture that alternately mixes along the time and feature dimensions, capturing both temporal dynamics and cross-variable interactions.
    \item TimeMixer~\cite{timemixer}: A multiscale MLP architecture that employs decomposable mixing blocks to model temporal patterns at multiple resolutions.
\end{itemize}

\begin{figure}[htbp]
\centering
\includegraphics[width=0.85\linewidth]{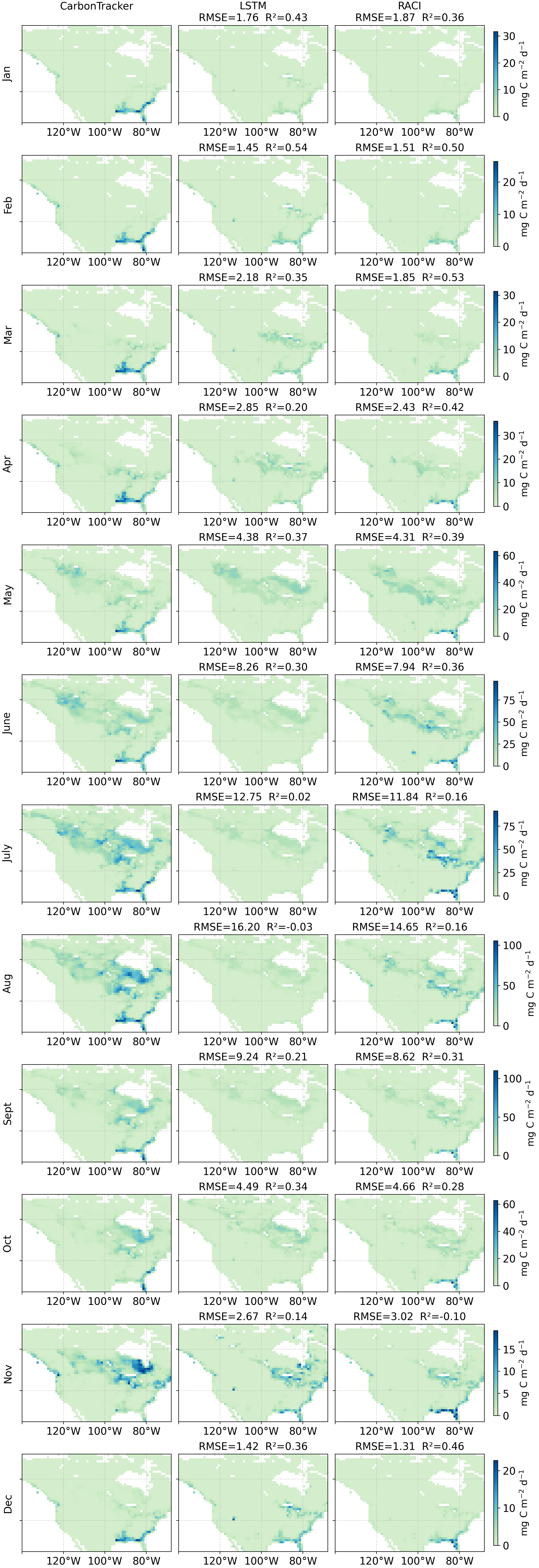}
\caption{Monthly CH$_4$ flux maps over North America in 2018 from CarbonTracker, LSTM, and \mthd prediction.}
\label{fig:carbontracker_monthly}
\end{figure}

\subsection{Implementation Details}
\label{sec:appendix_implement}

All experiments are implemented using PyTorch 2.11 and run on a single NVIDIA GTX 3080 GPU, using the ADAM optimizer with an initial learning rate of 0.001. All models are set with 3 layers and hidden state dimensions of 32, and Transformer-based models use 4 attention heads. We use mean squared error (MSE) as the training loss. For the simulation experiments, all models are trained with a batch size of 32, whereas for the observation-based experiments we use a smaller batch size of 4, keeping the batch size consistent across all baselines within each setting. For yearly retrieval, we first apply PCA to the yearly embeddings and retain the top four principal components. Cosine similarities are computed in this reduced space, and only candidate sequences whose cosine similarity with the target exceeds 0.99 are included in the retrieval data pool; these settings are fixed across all experiments and not tuned per site. For all baseline implementations, we adopt the official open-source code from the Time-Series Library (TSLib)\footnote{\url{https://github.com/thuml/Time-Series-Library}} and use the recommended configurations to ensure a fair comparison~\cite{wu2023timesnet,wang2024tssurvey}.

\subsection{Additional Experimental Results}
\label{sec:appendix_exp}

\subsubsection{FLUXNET-CH$_4$ site-level time-series comparison}
\label{sec:appendix_fluxnet_ts}

Figure~\ref{fig:FLUXNET_ts} illustrates daily CH$_4$ emission time series at five representative FLUXNET-CH$_4$ sites for individual evaluation years. Each panel compares observations with the LSTM baseline and \mthd. Across sites, the LSTM baseline often produces a smoothed seasonal cycle and converges toward intermediate flux levels, especially for years with large intra-annual amplitude. As a result, high-emission episodes and sharp transitions tend to be damped. In contrast, \mthd more closely follows both the timing and magnitude of seasonal peaks while still preserving low-emission periods, thereby mitigating the regression-to-the-mean behavior observed in the LSTM for these examples. Because extreme CH$_4$ emission events contribute disproportionately to annual and regional budgets, being able to represent such peaks more faithfully is particularly important in practice; these qualitative cases suggest that \mthd can better preserve episodic high emissions compared with the baseline LSTM.

\subsubsection{CarbonTracker-based CH$_4$ flux maps over North America}
\label{sec:appendix_carbontracker}

In the main text (Figure~\ref{fig:carbontracker_mean}), we present annual mean CH$_4$ emissions over North America in 2018 from CarbonTracker, the LSTM baseline, and \mthd. Figure~\ref{fig:carbontracker_monthly} extends this comparison to monthly CH$_4$ flux maps. Consistent with the site-level FLUXNET-CH$_4$ time-series results, the LSTM baseline tends to predict flux that are close to a climatological mean. During winter months, when overall emissions are low, this behavior can still yield reasonable agreement with CarbonTracker and, in some cases, slightly smaller apparent biases than \mthd. However, during the summer peak season, the LSTM substantially underestimates regional emission maxima and smooths out strong temporal variability, leading to muted seasonal amplitudes in both space and time. In contrast, \mthd better preserves high-emission regions and their seasonal evolution, including wetland-dominated areas such as the U.S. Gulf Coast, while still maintaining plausible wintertime flux levels.

\begin{figure*}[htbp]
    \centering
    \begin{subfigure}{0.43\textwidth}
        \centering
        \includegraphics[width=\linewidth]{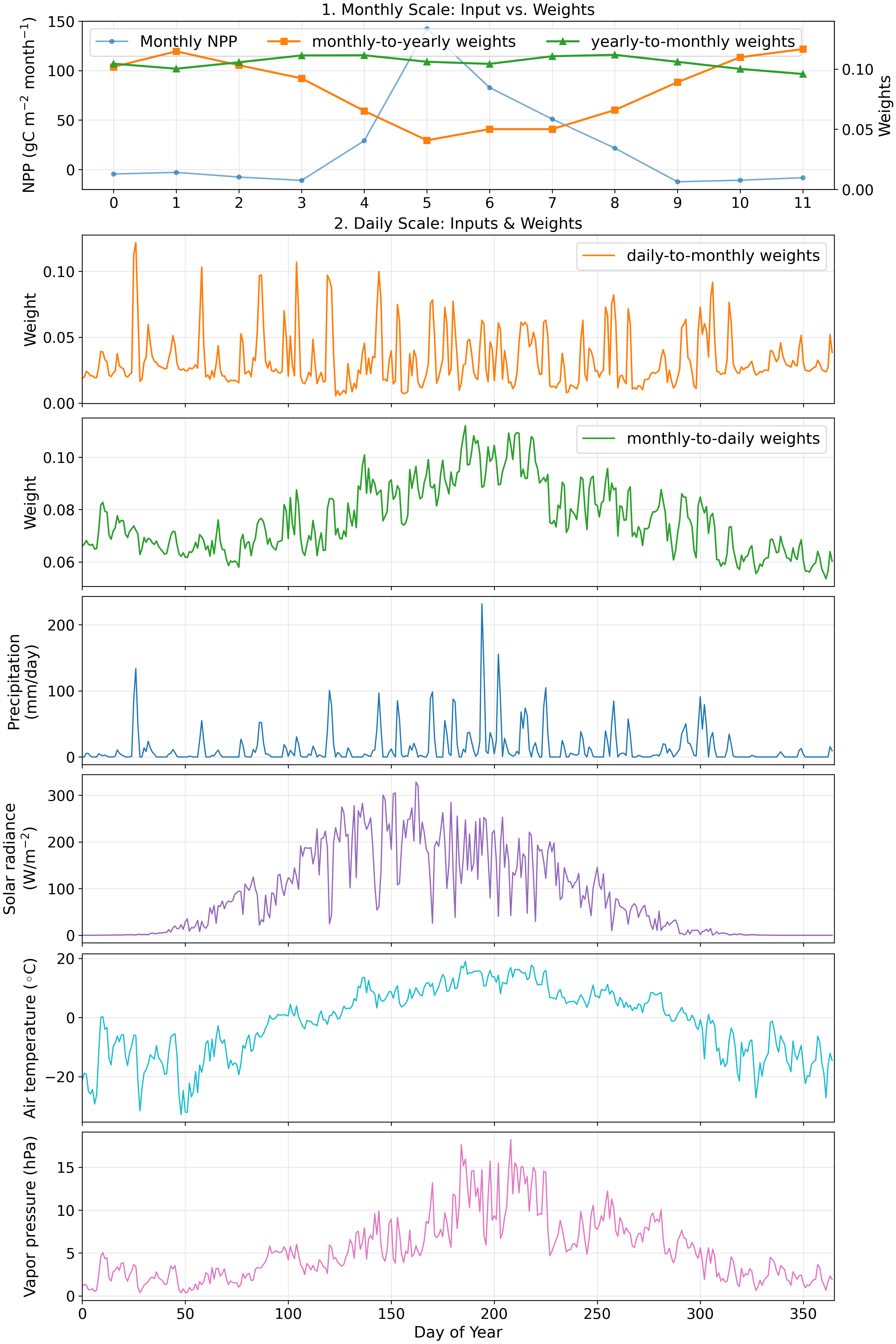}
        \caption{FI.Lom site} 
    \end{subfigure}
    \hspace{0.6cm}
    \begin{subfigure}{0.43\textwidth}
        \centering
        \includegraphics[width=\linewidth]{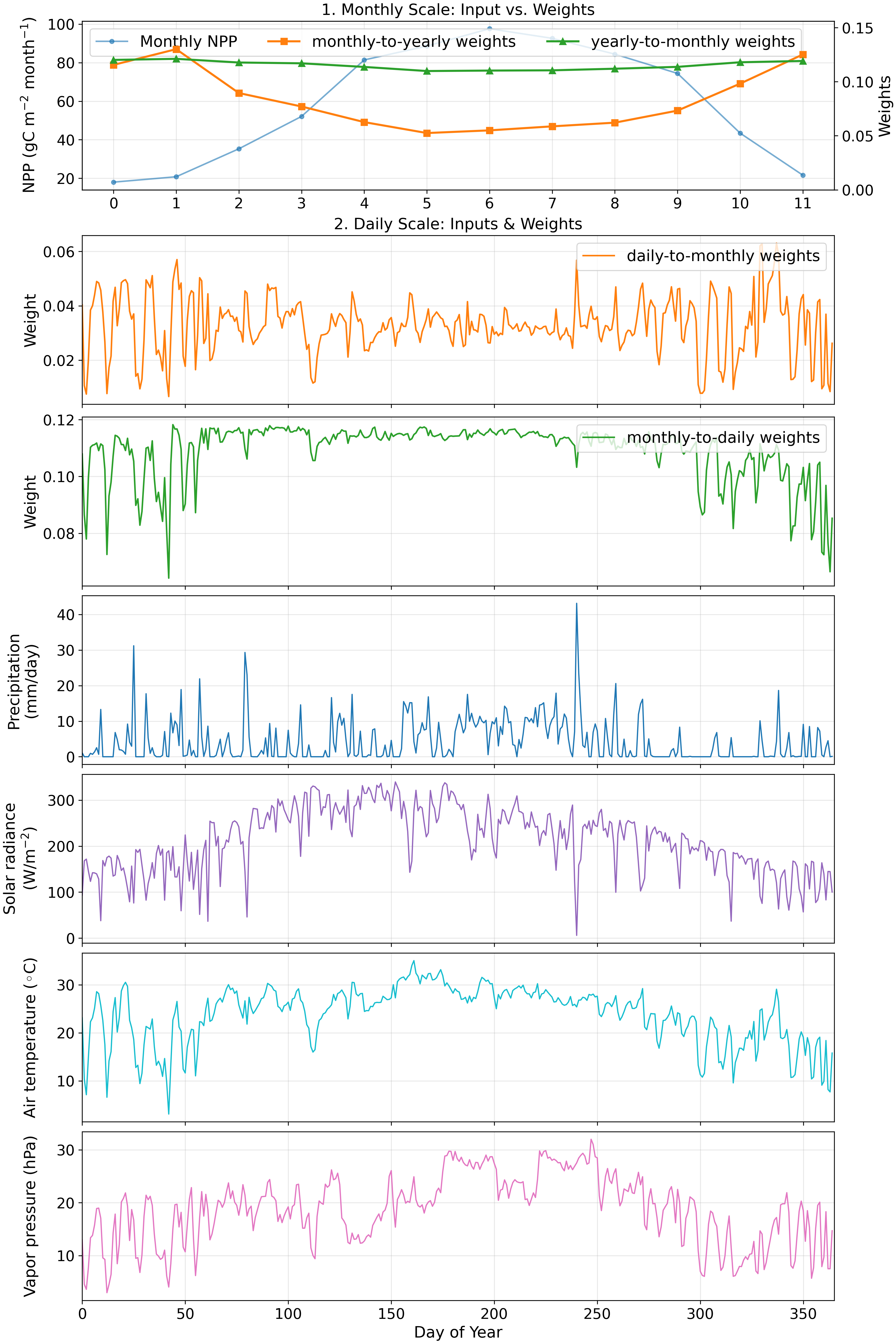}
        \caption{US.LA1 site}
    \end{subfigure}
    \vspace{-0.1in}
    \caption{Temporal attention patterns learned by \mthd for CH$_4$ prediction. }
    \label{fig:temporal_attention}
\end{figure*}

\begin{figure*}[htbp]
    \centering
    \begin{subfigure}{0.49\textwidth}
        \centering
        \includegraphics[width=\linewidth]{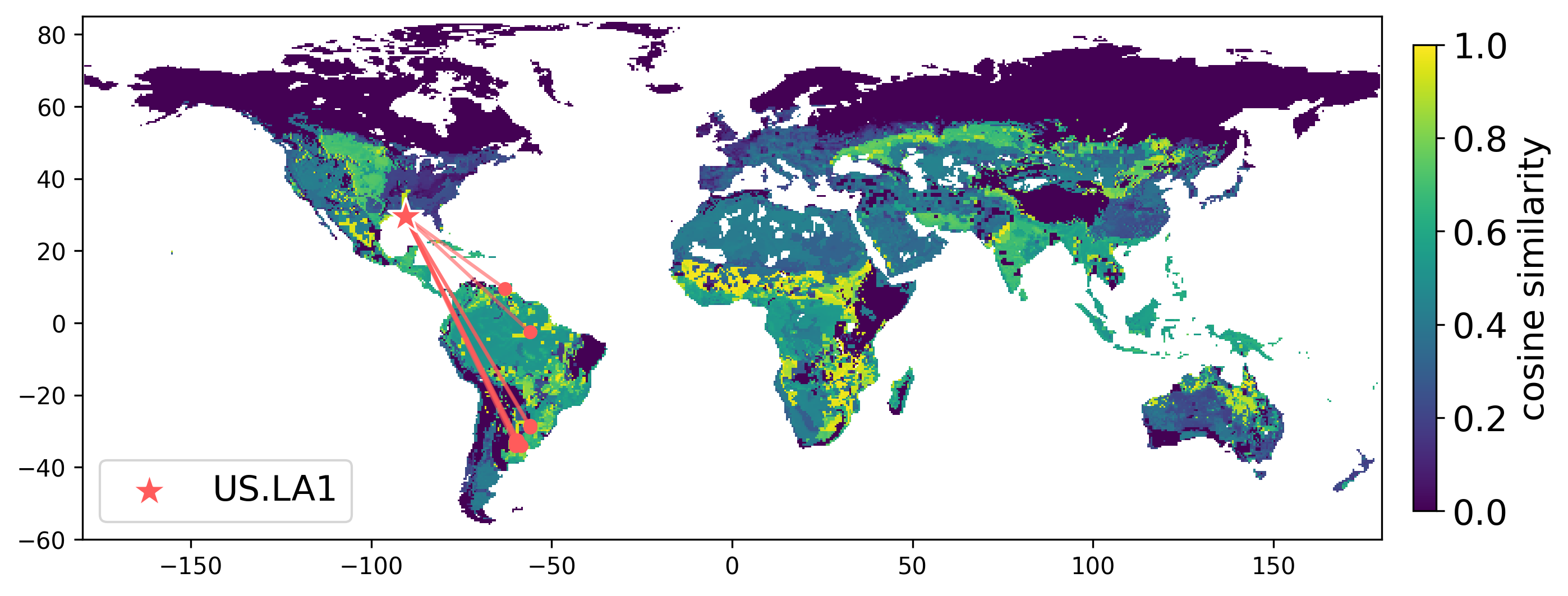}
        \label{fig:US.LA1}
    \end{subfigure}
    \hfill
    \begin{subfigure}{0.49\textwidth}
        \centering
        \includegraphics[width=\linewidth]{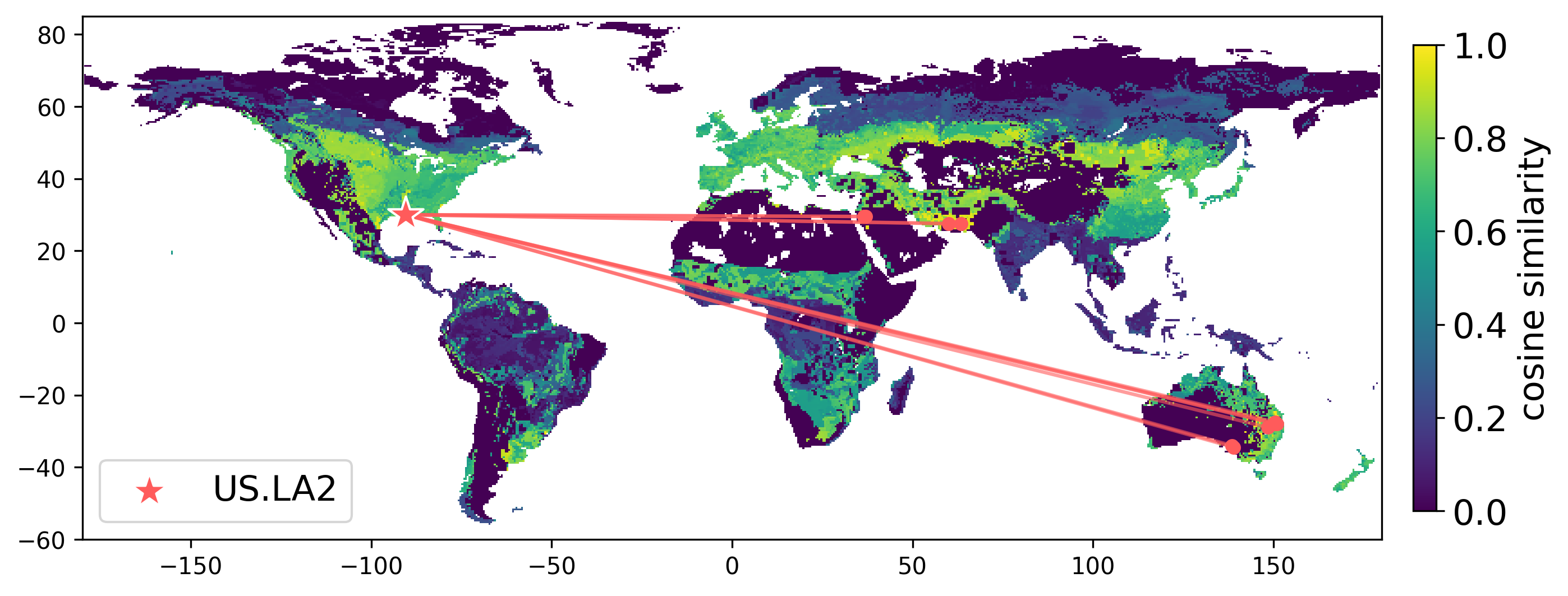}
        \label{fig:US.LA2}
    \end{subfigure}
    \vspace{-0.2in}
    \caption{Yearly spatial retrieval similarity maps and retrieved locations for two adjacent FLUXNET-CH$_4$ sites in the Mississippi River Delta (US-LA1, left; US-LA2, right).}
    \label{fig:la_retrieval}
\end{figure*}

\subsubsection{Visualization of temporal attention}
\label{sec:appendix_temporal_attention}

Figure~\ref{fig:temporal_attention} provides the visualizations referred to in Section~\ref{sec:temporal_attention_main}. For two representative FLUXNET-CH$_4$ sites (FI-Lom and US-LA1), the top panels show monthly NPP inputs together with attention weights between monthly and yearly steps in the aggregation and propagation modules. The subsequent panels show attention weights between daily and monthly steps in the aggregation and propagation modules, alongside key meteorological drivers (precipitation, solar radiation, air temperature, and vapor pressure). These plots illustrate the patterns discussed in the main text: yearly descriptors emphasize seasons with strong contrast, daily aggregation focuses on whichever driver limits local flux, and propagation gates amplify regime information primarily during methane-favourable periods.

\subsubsection{Visualization of spatial retrieval}
\label{sec:appendix_spatial_retrieval}

Figure~\ref{fig:la_retrieval} complements the discussion in Section~\ref{sec:yearly_retrieval_main} by visualizing yearly spatial similarity maps and retrieved locations for two adjacent sites in the Mississippi River Delta (US-LA1 and US-LA2). For each target site (red star), we first compute yearly embeddings, apply PCA to retain the top four components, and then evaluate cosine similarity between the target and all grid cells in the auxiliary domain. The background color shows this cosine similarity field, and the red lines highlight the ten most similar grid cells with cosine similarity greater than 0.99, i.e., the locations that enter the retrieval data pool. 

Although US-LA1 and US-LA2 are geographically close and share a similar climate, their yearly retrieval neighborhoods are very different: the former mainly retrieves other wetland systems, whereas the latter retrieves managed agroecosystems. This indicates that \mthd’s yearly embeddings organize sites by inferred hydrological and ecological regimes instead of simple geographic distance or coarse climate similarity.

\subsection{Parameter Sensitivity Analysis}

We further evaluate the sensitivity of \mthd to two key inference-time hyperparameters: (i) the similarity threshold $\tau$ for entering the retrieval pool, varied in $\{0.95, 0.97, 0.99\}$, and (ii) the number of principal components $k$ used in the PCA-based embedding, varied in $\{3, 4, 5\}$. 
We adopt a controlled variable approach to isolate the impact of each factor: we fix the default setting ($\tau=0.99, k=4$) and vary one hyperparameter at a time. 
As reported in Table~\ref{tab:sensitivity}, the performance remains highly stable across these variations. 
This variation range is smaller than the stochastic variability observed when retraining models from different random initializations (as shown in Table~\ref{tab:ch4_table}). 
These results confirm that \mthd is robust to reasonable choices of hyperparameters and does not require fine-grained tuning at inference time.

\begin{table}[h]
    \centering
    \caption{Sensitivity analysis of RACI on TEM-MDM under different hyperparameter settings.}
    \vspace{-0.1in}
    \label{tab:sensitivity}
    \begin{subtable}{0.48\linewidth}
        \centering
        \caption{Varying Similarity Threshold ($\tau$) with fixed PCA$=4$}
        \label{tab:sens_threshold}
        \resizebox{\linewidth}{!}{
            \begin{tabular}{cc|cc}
            \toprule
            Threshold ($\tau$) & PCA & RMSE & $R^2$ \\
            \midrule
            0.95 & 4 & 2.13 & 0.97 \\
            0.97 & 4 & 2.10 & 0.97 \\
            \textbf{0.99} & \textbf{4} & \textbf{2.08} & \textbf{0.97} \\
            \bottomrule
            \end{tabular}
        }
    \end{subtable}
    \hfill
    \begin{subtable}{0.48\linewidth}
        \centering
        \caption{Varying PCA Components with fixed Threshold $\tau=0.99$}
        \label{tab:sens_pca}
        \resizebox{\linewidth}{!}{
            \begin{tabular}{cc|cc}
            \toprule
            Threshold ($\tau$) & PCA & RMSE & $R^2$ \\
            \midrule
            0.99 & 3 & 2.11 & 0.97 \\
            \textbf{0.99} & \textbf{4} & \textbf{2.08} & \textbf{0.97} \\
            0.99 & 5 & 2.10 & 0.97 \\
            \bottomrule
            \end{tabular}
        }
    \end{subtable}
\end{table}

\subsection{Uncertainty Quantification}

\begin{figure}[htbp]
\centering
\includegraphics[width=\linewidth]{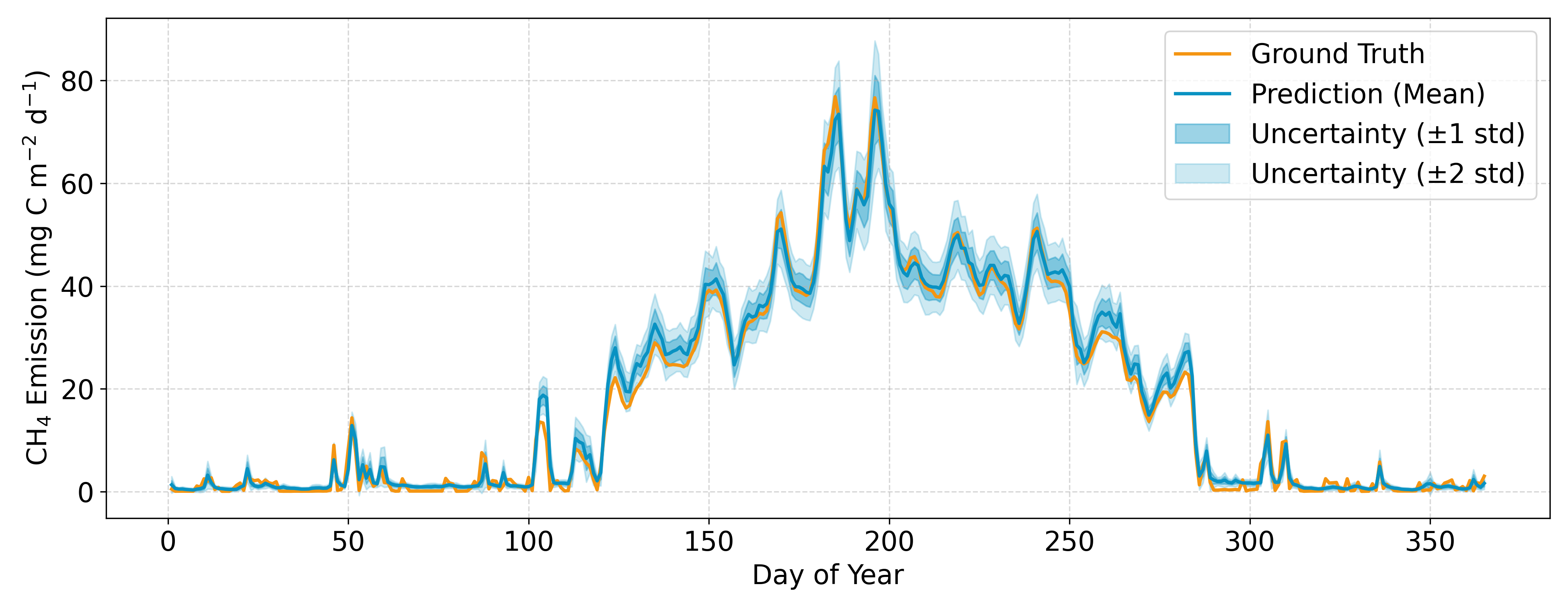}
\vspace{-0.2in}
\caption{Example site–year from TEM-MDM showing ground-truth flux, \mthd prediction, and MC Dropout uncertainty bands (mean $\pm$ 1 std and $\pm$ 2 std).}
\label{fig:uncertainty}
\end{figure}

We assess the reliability of our predictions using Monte Carlo (MC) Dropout, keeping dropout active at test time to generate an ensemble of stochastic forward passes. We use a dropout rate of $p=0.1$ in all dropout-compatible layers (LSTM units, multi-head attention, and fully connected layers) and perform $T=50$ forward passes per test sample, taking the mean as the prediction and the standard deviation as an estimate of epistemic uncertainty. Averaged over all test site–years, the ratio between predictive standard deviation and mean absolute flux is 0.1139 for LSTM and 0.1103 for \mthd, indicating that \mthd does not inflate uncertainty relative to a strong baseline and in fact yields slightly smaller epistemic spread. In the time-series plots (Figure~\ref{fig:uncertainty}), the mean $\pm$ one-standard-deviation bands widen around sharp emission peaks and other hard-to-fit periods, while remaining narrow over more stationary segments, suggesting that the inferred uncertainty concentrates on regions of reduced predictive skill rather than being uniformly large.

\subsection{Limitations and Future Directions}
A key limitation of \mthd lies in its reliance on an auxiliary data pool for spatial contextual retrieval. Computationally, retrieval over this pool introduces a dependence on both the number of target site-years and the number of candidates, which can become costly if the pool is naively scaled to very large domains. Methodologically, the quality and diversity of the pool determine how well the retrieved context covers the regimes encountered at inference time. When a new site-year falls into a regime that is poorly represented in the pool, the retrieval module may return no candidates above the similarity threshold. In such cases, we set the retrieved trajectory to zero so that \mthd gracefully degenerates to a standard global predictor, rather than forcing mismatched contextual information into the prediction. 

These limitations point to a common avenue for improvement: designing smaller yet higher-quality auxiliary pools. Promising directions include curating compact sets of regime prototypes, enforcing diversity in background conditioners, and augmenting underrepresented regimes with data-driven synthetic trajectories or additional simulators. Such developments would reduce computational cost while improving coverage of rare regimes, further strengthening the robustness and scalability of role-aware retrieval.

\end{document}